\documentclass[journal,final,twoside]{IEEEtran}

\usepackage[pdftex]{graphicx}
\graphicspath{{figures/}}
\usepackage[utf8]{inputenc}

\usepackage{soul}
\usepackage{cite}

\usepackage{array}
\newcolumntype{C}{>{\centering\arraybackslash}b{1.6cm}}
\newcolumntype{B}{>{\raggedright\arraybackslash}b{2.7cm}}

\usepackage{rotating}
\usepackage{pdfpages}

\title{Language Bootstrapping: Learning Word Meanings From Perception--Action Association}

\author{Giampiero~Salvi,~
        Luis~Montesano,~
        Alexandre~Bernardino,~\IEEEmembership{Member,~IEEE,}
        José~Santos-Victor,~\IEEEmembership{Member,~IEEE}% <-this % stops a space		
\thanks{Manuscript received October 26, 2010; revised July 1, 2011; accepted October 6, 2011. Date of publication November 16, 2011; date of current version May 16, 2012. This work was supported in part by European Union New and emerging science and technologies Project 5010-Contact and in part by Fundação para a Ciência e Tecnologia (Institute for Systems and Robotics/Instituto Superior Técnico plurianual funding) through the Programa Operacional Sociedade de Conhecimento Program that includes European fund for regional development funds. This paper was recommended by Editor E. Santos, Jr.}
        \thanks{G. Salvi is with the Speech, Music and Hearing department at Kungliga Tekniska Högskolan (KTH), Stockholm, Sweden. giampi@kth.se}% <-this % stops a space
\thanks{L. Montesano is with the Computer Science Department, Universidad de Zaragoza, Spain. montesano@unizar.es}% <-this % stops a space
\thanks{A. Bernardino and J. Santos-Victor are with the Instituto de Sistemas e Robótica, Instituto Superior Técnico, Lisboa, Portugal. \{alex,jasv\}@isr.ist.utl.pt}% <-this % stops a space
\thanks{Color versions of one or more of the figures in this paper are available online at http://ieeexplore.ieee.org.}% <-this % stops a space
\thanks{Digital Object Identifier 10.1109/TSMCB.2011.2172420}
} % end \author

\ifCLASSOPTIONpeerreview
\else
\fi
\IEEEpubid{1083-4419/\$26.00~\copyright~2011 IEEE}

\begin{document}
\setcounter{page}{660}
\maketitle
\IEEEpeerreviewmaketitle

\begin{abstract}
We address the problem of bootstrapping language acquisition for an artificial system similarly to what is observed in experiments with human infants. Our method works by associating meanings to words in manipulation tasks, as a robot interacts with objects and listens to verbal descriptions of the interactions. The model is based on an affordance network, i.e., a mapping between robot actions, robot perceptions and the perceived effects of these actions upon objects. We extend the affordance model to incorporate spoken words, which allows us to ground the verbal symbols to the execution of actions and the perception of the environment.
The model takes verbal descriptions of a task as the input, and uses temporal co-occurrence to create links between speech utterances and the involved objects, actions and effects. We show that the robot is able form useful word-to-meaning associations, even without considering grammatical structure in the learning process and in the presence of recognition errors. These word-to-meaning associations are embedded in the robot’s own understanding of its actions. Thus, they can be directly used to instruct the robot to perform tasks and also allow to incorporate context in the speech recognition task.
We believe that the encouraging results with our approach may afford robots with a capacity to acquire language descriptors in their operation's environment as well as to shed some light as to how this challenging process develops with human infants.
\end{abstract}


\section{Introduction}
\label{sec:introduction}

\IEEEPARstart{T}{o interact} with humans, a robot needs to communicate with people and  understand their needs and intentions. By far the most natural way for a human to communicate is language. This paper deals with the acquisition by a robot of language capabilities linked to manipulation tasks. 
Our approach draws inspiration from infant cross situational word learning theories that suggest that infant learning is an iterative process involving multiple strategies \cite{Akhtar99,smith2011}. It occurs in an incremental way (from simple words to more complex structures) and involves multiple tasks such as word segmentation, speech production, and meaning discovery. Furthermore, it is highly coupled with other learning processes such as manipulation, for instance, in mother infant interaction schemes \cite{Lacerda04}.

\IEEEpubidadjcol % to avoid \IEEEpubid overlap with text

Out of the multiple aspects of language acquisition, this paper focuses on the ability to discover the meaning of words through human-robot interaction.
We adopt a developmental robotics approach \cite{weng98developmental,lungarella03develsurvey} to tackle the language acquisition problem.  In particular, we consider the developmental framework of \cite{Montesano08} where the robot first explores its sensory-motor capabilities. Then, it interacts with objects and learns their affordances, i.e. relations between actions and effects. The affordance model uses a Bayesian network to capture the statistical dependencies among a set of robot basic manipulation actions (e.g. grasp or tap), object features and the observed effects by means of statistical learning techniques exploiting the co-occurrence of stimuli in the sensory patterns.

The main contribution of the paper is the inclusion in the affordance model \cite{Montesano08} of verbal descriptions of the robot activities, provided by a human. The affordance model encodes possible meanings in terms of the relation between actions, object properties and effects grounded in the robot experience. The extended model exploits temporal co-occurrence to associate speech segments to these {\em affordance} meanings. Despite we do not use any social cues or the number and order of words, the model provides the robot with the means to learn and refine the meaning of words in such a way that it will develop a rough understanding of speech based on its own experience. 

Our model has been evaluated using a humanoid torso able to perform simple manipulation tasks and to recognize words from a basic dictionary. We show that simply measuring the frequencies of words with respect to a self-constructed model of the world, the affordance network, is sufficient to provide information about the meaning of these utterances even without considering prior semantic knowledge or grammatical analysis. By embedding the learning into the robot's own task representation, it is possible to derive links between words such as nouns, verbs and adjectives and the properties of the objects, actions and effects. We also show how the model can be directly used to instruct the robot and to provide contextual information to the speech recognition system.

Although the paper follows the approach in \cite{KrunicEtAl2009ICRA}, the results are based on new data and on a different treatment of the data. In particular, the design of the sentence material describing the affordance experiments and the speech recordings and recognition of the material have been improved. In addition to this, we have analyzed the impact of the model in ambiguous situations by comparing the robot answers to human answers.

The rest of the paper is organized as follows. After discussing related work, Section~\ref{sec:approach} briefly describes, through our particular robotic setup, the problem and the general approach to be taken in the learning and exploitation phases of the word-concept association problem. Section~\ref{sec:method} presents the language and manipulation task model and the algorithms used to learn and make inferences. In Section~\ref{sec:experiments} we describe the experiments and provide some details on the speech recognition methods employed. Results are presented in Section~\ref{sec:results} and finally, in Section~\ref{sec:conclusions}, we conclude our work and present ideas for future developments.

\section{Related Work}
Computational models for cross situational word learning have only been studied recently. One of the earliest works is the one by Siskind \cite{Siskind1996} who proposes a mathematical model and algorithms for solving an approximation of the lexical-acquisition task faced by children. The paper includes computational experiments, using a rule based logical inference system, which shows that the acquisition of word-to-meaning mappings can be performed by constraining the possible meanings of words given their context of use. They show that acquisition of word-to-meaning mappings might be possible without knowledge of syntax, word order or reference to properties of internal representations other than co-occurrence.  This has motivated a series of other research in cross-situational learning.
For instance, Frank, Goodman and Tenenbaum \cite{Frank08} presented a Bayesian model for cross-situational word-learning that learns a "word-meaning" lexicon relating objects to words. Their model explicitly deals with the fact that some words do not represent any object, e.g., a verb or an article. By modeling the speaker's intentions, they are also able to incorporate social cues typically used by humans. 

In the last years, there has been an effort to understand the language acquisition process during the early years of life of an infant. Analysis from the recordings of the first three years of life of a child suggest that caregivers fine tune their interaction in a way that can definitely shape the way language is acquired \cite{DebRoy09}.  

In order to develop natural human-robot interfaces, recent works have established bridges between language acquisition models, natural language processing techniques and robotic systems. One of the challenges arising from such a combination is that robots do not deal only with speech. As humans, they operate in a continuous world, perceive it and act on it. The multi-modal information may greatly help developing language skills, but also requires to consider the different nature of the information and their coordination. Two related recurrent topics in the literature for robot language acquisition are embodiment \cite{PfeiferBook} and symbol grounding \cite{Harnad90}. The former states that learning is shaped by the body. As a result, the internal representations of information tie together action and perception. A relevant example in the context of language is the affordance concept \cite{gibson79ecological}. More specifically, object affordances have been pointed out as a promising internal representation to capture the relations between objects, actions and consequences \cite{Montesano08}.  

On the other hand, language grounding links the symbolic nature of language with the sensory-motor experience of the robot. Most of the works, in this case, focus on associating names to objects through their perceptions. For instance, in \cite{Lopes07} the robot learns word-object associations through incremental one-class learning algorithms. The focus is on open-ended, long term learning.  Objects are represented using many different shape-based features and categories are simply represented by instances. 

Recent works have also addressed actions and their consequences. The work in \cite{Takamuku06} exploits object behavior (resulting effects of an action) to create object categories using reinforcement learning. Without considering learning, \cite{Mavridis06} proposed a layered grounded situation model comprised of three layers that go from the continuous to a symbolic representation and allows the robot to understand the current situation, reason about its own experience and make predictions. In \cite{Hsiao08}, affordances have been used to ground language by constructing object schemes. An object schema is a hand-coded description of the object in terms of potential interactions related to the object and allows to plan, predict or recognize according to them. 

Probably, one of the most interesting works, in our perspective, is the one presented in \cite{gs:YuAndBallard2004,Yu07}. Here a human subject was instrumented with devices to perceive its motor actions, speech discourse and the interacting objects (camera, data glove and microphone), and an automatic learning system was developed to associate phoneme sequences to the performed actions (verbs) and observed objects (nouns). Common phoneme patterns were discovered in the speech sequence by using an algorithm based on Dynamic Programming. These patterns were then clustered into similar groups using and agglomerative clustering algorithm in order to define wordlike symbols to associate to concepts. 

Finally, \cite{He07} proposed a self-organizing incremental neural network to associate words to object properties. The system uses fixed rules where the teacher provides the labels to specific objects via pointing. A single label-perception pair is used to create the model. Once nouns have been learned, the same procedure is used to learn verbs.

Our approach is similar to the one presented in \cite{Yu07} in the sense that we also consider the interaction between robot and object to be described by multiple sources of information (acoustic, visual and motor). However, due to the embodiment inherent to the robot, the latter has access to its own actions which removes the need to estimate the action from video-recorded sequences.  Also, the robot interacts with a single object at a time and, consequently, there is no need for a mechanism to infer attention.
Finally, we assume that the robot has already learned the acoustics of a set of words and is able to recover them from the auditory input. We leave out of the current study the problem of learning the words from sequences of acoustic classes as in \cite{gs:YuAndBallard2004, gs:StoutenEtAl2008} and learning the acoustic classes from the speech signal as in \cite{gs:Salvi2005a}. In spite of these simplifying assumptions, in this study, objects are represented by their features (shape, color, size) rather than by their category, thus allowing for a more flexible description than some previous approaches. As a result, our model automatically incorporates adjectives (object properties). The affordance model includes the description of effects (outcomes of actions), therefore addressing the acquisition of concepts related to behaviors (e.g ``the ball is moving'', ``the box is still'').

\section{Approach}
\label{sec:approach}
In this section, we provide an overview of the full system. As mentioned before, we assume that the robot is at a developmental stage where basic manipulation skills have already been learned up to a maturity level that includes a model of the results of these actions on the environment (see \cite{Montesano08} for further details). In order to make the presentation less abstract, we describe the particular robotic setup used in the experiments and the skills already present in the system.

\begin{figure}
 \centering
\includegraphics[width=0.8\columnwidth]{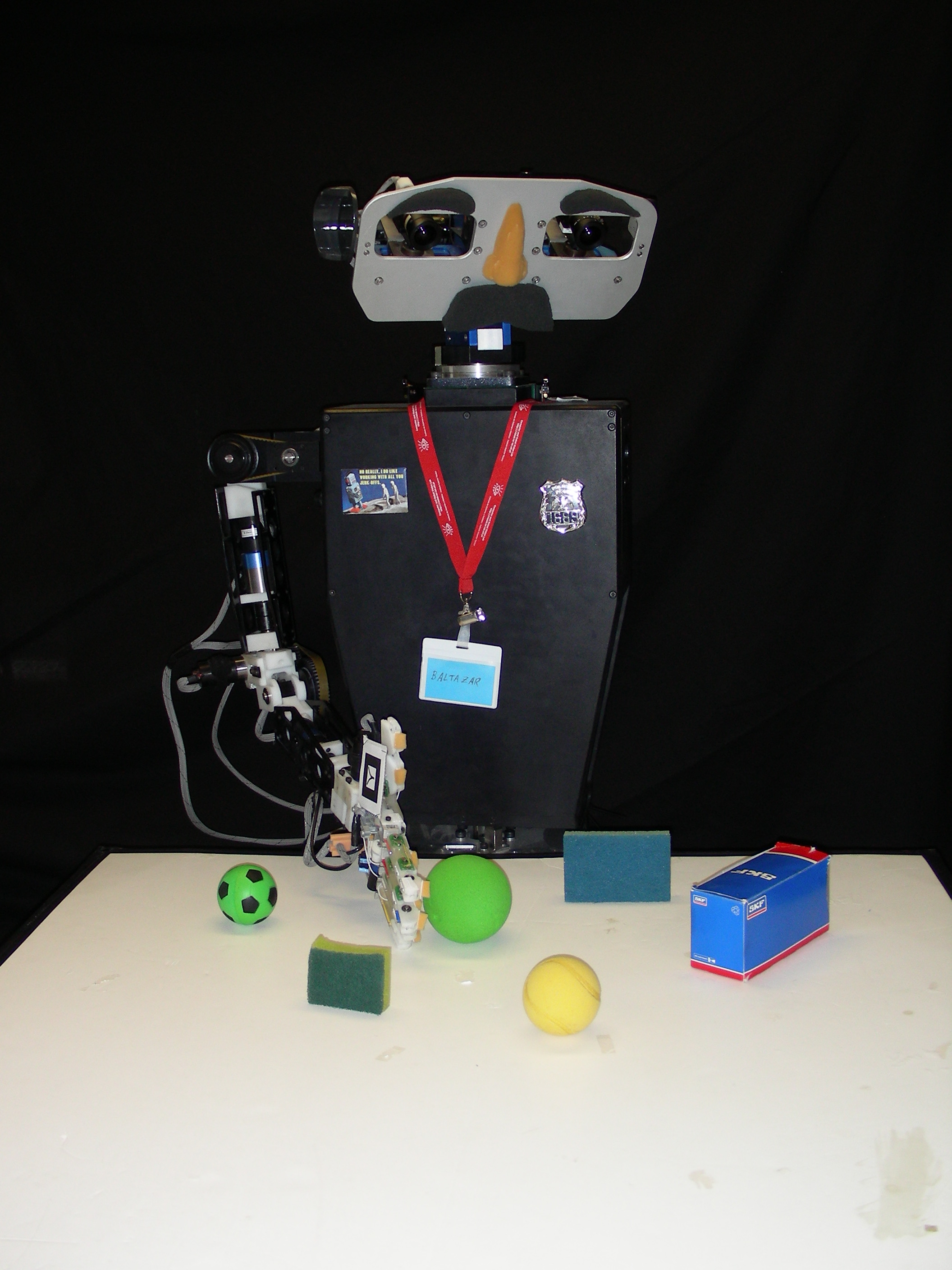}
\caption{Baltazar, the humanoid torso used in the experiments.}
\label{fig:balta}
\end{figure}

\subsection{Robot skills and developmental stage}
We used Baltazar, a $14$ degrees of freedom humanoid torso composed by a binocular head and an arm (see Figure~\ref{fig:balta}).

The robot is equipped with the skills required to perform a set of simple manipulation actions denoted by $a_i$ on a number of objects. In our particular experiments we consider the actions \emph{grasp}, \emph{tap} and \emph{touch}. In addition to this, its perception system allows it to detect objects placed in front of it and extract information about them. More precisely, it extracts from the raw sensory data some continuous visual descriptors of its color, size and shape. These continuous descriptors are clustered in an unsupervised way to form symbolic descriptions (discrete labels) of the object characteristics. These are represented in a feature vector $\mathbf{f}=(f_1,f_2,f_3)$, where $f_1$, $f_2$ and $f_3$ are, respectively, the color, size and shape discrete feature labels. After performing the action, the robot detects and categorizes the effects produced by its actions. Effects are mainly identified as changes in the perception such as the object velocity ($e_1$), the velocity of the robot's own hand ($e_2$), the relative velocity between object and hand ($e_3$) and the persistent activation of the contact sensors in the hand ($e_4$). This information is also obtained from unsupervised clustering of corresponding continuous sensory data, and stored in feature vector $\mathbf{e}=(e_1, e_2, e_3, e_4)$.

Once these basic action-perception skills have been acquired, the robot undergoes a self-exploratory training period that allows it to establish relations between the actions $a$, the object features $\mathbf{f}$ and the  effects $\mathbf{e}$.
%\footnote{This is not   strictly necessary in the model presented in the next   section. However, in order to test the expressiveness of the method   we made this assumption.}. 
This model captures the world behavior under the robot actions. It is important to note that the model includes the  notion of consequences\footnote{One should be always careful about causality inference. However, under certain constraints one can at least guess about induced statistical dependencies \cite{Pearl00}.} and, up to a certain extent, an implicit narrative structure of the execution of an action upon an object.

The robot is also equipped with audio perception capabilities that allow it to recover an uncertain list of words ($\{w_i\}$) from the raw speech signal ($s$) based on a previously trained speech recognizer.

\subsection{Incorporating speech}
Based on the existing cognitive capabilities of the robot, described above, we aim at exploiting the co-occurrence of verbal descriptions and simple manipulation tasks to associate meanings and words. Our approach is the following:
\begin{enumerate} 
\item During the execution of an action ($a$), the robot listens to the users speech and recognizes some words ($\{w_i\}$). The words are stored in a bag of words model, i.e. an unordered set where multiple occurrences are merged. 
\item These recognized words are correlated with the concepts of actions ($a$), object features ($\mathbf{f})$ and effects ($\mathbf{e}$) present in the world. Our objective is to learn the correct relationships between the word descriptions and the previous manipulation model through a series of robot-human interaction experiments. These relations implicitly encode word-meaning associations grounded to the robot's own experience.
\end{enumerate}

\begin{figure}
 \centering
\includegraphics[width=0.95\columnwidth]{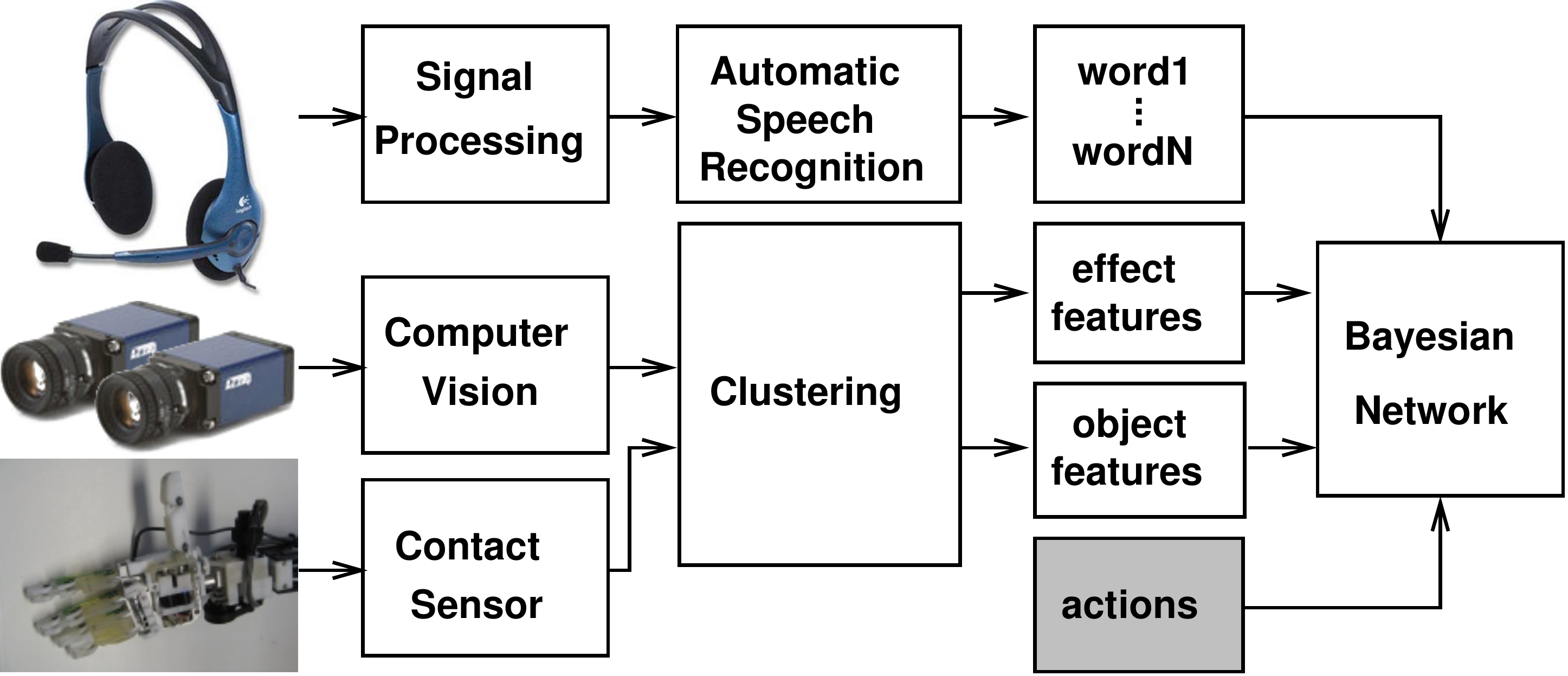}
\caption{Overview of the setup.}
\label{fig:setup}
\end{figure}

We model this problem in a Bayesian probabilistic framework where the actions $A$, defined over the set $\mathcal{A}=\{a_i\}$, object properties $F$, over $\mathcal{F}=\{f_i\}$ and effects $E$, over $\mathcal{E}=\{e_i\}$ are random variables. We will denote $X=\{A, F, E\}$ the state of the world as experienced by the robot. The joint probability $p(X)$ encodes the basic world behavior grounded by the robot through interaction with the environment. The verbal descriptions are denoted by the set of words $W=\{w_i\}$. Figure~\ref{fig:setup} illustrates all the information fed to the learning algorithm.

If we consider the world concepts or meanings being encoded by $X$, then, to learn the relationships between words and concepts, we estimate the joint probability distribution $p(X,W)$ of actions, object features, effects, and words in the speech sequence. Once good estimates of this function are obtained, we can use it for many purposes, for example:
\begin{itemize}
\item to compute associations between words and concepts, by estimating the structure of the joint pdf $p(X,W)$;
\item to plan the robot actions given verbal instructions from the user in a given context, through $p(A, F \mid W)$;
\item to provide context to the speech recognizer by computing $p(W \mid X)$.
\end{itemize}

\section{Model - Algorithms}
\label{sec:method}
In this section, we present the model and methods used to learn the relations between words and the robot's own understanding of the world. Our starting point is the affordance model presented in \cite{Montesano08}. This model uses a discrete Bayesian network to encode the relations between the actions, object features and the resulting effects. The robot learns the network from self-experimentation with the environment and the resulting model captures the statistical dependencies among actions, object features and the consequences of the actions.

\subsection{Learning Word-to-Meaning Associations}
\begin{figure}
\centering
 \includegraphics[width=0.5\columnwidth]{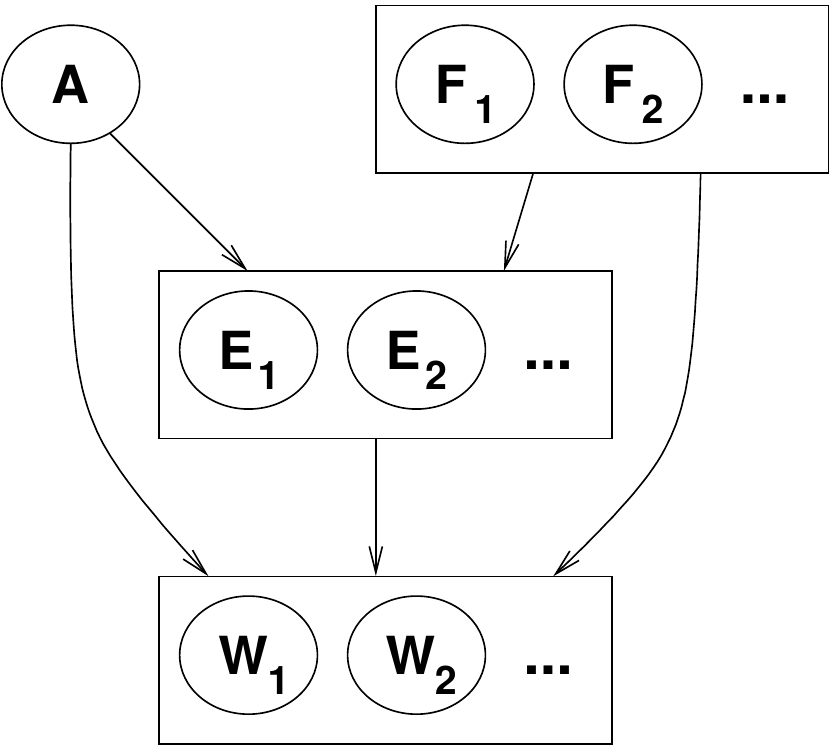}
\caption{Graphical representation of the model. The affordance network is represented by three different sets of variables: actions ($A$), object features ($F_i$) and effects ($E_i$). Each word $w_i$ may depend on any subset of $A$, $F_i$ and $E_i$.}
\label{fig:model}
\end{figure}

Here we explain how the model of \cite{Montesano08} is extended to include also information about the words describing a given experience. Recall that $X$ denotes the set of (discrete) variables representing the affordance network. For each word in $W$, let $w_i$ represent a binary random variable. A value $w_i=1$ indicates the presence of this word, while $w_i=0$ indicates the absence of this word in the description. We impose the following factorization over the joint distribution on $X$ and $W$
\begin{eqnarray}
 P(X,W) & = & \prod_{w_i \in W} p(w_i \mid X_{w_i} ) p(X)
\label{eq:model}
\end{eqnarray}
where $X_{w_i}$ is the subset of nodes of $X$ that are parents of word $w_i$. The model implies that the set of words describing a particular experience depends on the experience itself\footnote{This point requires a careful treatment when dealing with baby language learning and, usually, explicit attention methods are required to constrain the relations between words and the meanings they refer to.}. On the other hand, the probability of the affordance network is independent of the words and, therefore, is equal to the one in \cite{Montesano08}. Figure~\ref{fig:model} illustrates the generic model proposed in the paper. 
On one hand, effects may depend on the object features and the action applied upon the object. On the other hand, meaning of words is encoded in the dependencies of words on the affordance network. In other words, the affordance network provides the set of possible meanings for the words grounded on the own robot experience with the environment.  

A strong simplifying assumption of our model is representing phrases and sentences as an unordered set of words, disregarding grammar, word order, and repetitions. This is actually known as the \emph{bag of words} assumption and is widely used, for instance, in document classification \cite{JordanDPsDocClass}, and information retrieval. Furthermore, we assume words in the collection are mutually independent. Given a network structure, i.e. the set of $X_{w_i}$ per each word $w_i$, our model simply computes the frequency of such a word for each configuration of the parents.

The most challenging part of the problem is to select, based on the data, which affordance nodes actually are related to each of the words. This is basically a model selection problem and has been widely studied in the machine learning literature in the context of graphical models and Bayesian networks (see \cite{Hec98} for a review). As mentioned above, the model of Eq. \ref{eq:model} does not consider relations among the different words. Therefore, we explore possible dependencies between each word with each affordance node using a simple greedy approach known as K2 algorithm \cite{Cooper92} to select the most likely graph given a set of training data $D=\{(d_i)\}$. Each example $d_i=\{X_i, W_i\}$, i.e. it is a pair of a network configuration $X_i$ and a verbal description $W_i$.

\subsection{Exploiting Cross-Modal Associations}
After the model has been learned, we can use it for several different tasks. Let us briefly describe some inference queries that can be solved by our model. As mentioned in Section~\ref{sec:approach}, the network allows to perform several speech based robot-human interactions.

First, the robot can be instructed to perform a task. This corresponds to recovering the (set of) action(s) given the words $W_s$ recognized from the operator's speech signal, e.g. $p(A \mid W_s)$. When dealing with a particular context, i.e. a set of potential objects to interact with, the robot may maximize:
\begin{eqnarray}
 <a^*,o^*> & = & \arg max_{a_i, o_i \in O_s} p(a_i, F_{o_i} \mid W_S) \label{eq:selectactionobject}\\
 & \propto & \prod_{w_i \in W_s} p( w_i \mid a_i, F_{o_i} ) p (a_i, F_{o_i})
\end{eqnarray}
where $O_s$ is the set of objects detected by the robot and $F_{o_i}$ the features associated to object $o_i$.

Assuming that we have non informative priors over the actions and objects, the robot seeks to select the action and object pair that maximizes the probability of $W_s$, i.e. it is more ``consistent'' with the verbal instruction. Alternatively, the robot may compute the $k$-best pairs. Notice that the model allows for incomplete and flexible inputs: The verbal input may specify object properties in a possibly ambiguous way, and it may specify an effect we want to obtain rather than explicitly an action we want the robot to perform (e.g. ``move the ball'', rather than ``grasp'' or ``tap'').

Second, the proposed model also allows to use context to improve recognition. Consider the case where the recognizer provides a list of $m$ possible sets of words $W^j_s$, $j \in 1..m$. The robot can perform the same operation as before to decide what set of words is the most probable or rank them according to their posterior probabilities. In other words, one can combine the confidence of the recognizer on each sentence with the context information to select among the possible sets of words by computing for each $W^i_s$
\begin{eqnarray}
  p(W^j_S \mid X) \propto & \left[\prod_{w_i \in W^j_s} p( w_i \mid X)\right] p(W^j_S) \label{eq:combinerecog}
\end{eqnarray}
where $p(W^j_S)$ is the probability of sequence $j$ according to the recognizer. 
\section{Experiments}
\label{sec:experiments}

\subsection{Affordance Data} \label{sec:affdata}
The manipulation experiments used to train the network are the same as in \cite{Montesano08}. Actions were discrete by design (touch, tap and grasp). Objects were described based on continuous descriptors for three object features: shape, color and size. Shape was encoded in six region-based descriptors, convexity, eccentricity,  compactness, roundness, squareness computed directly from the segmented image of the object. Size was extracted from the two axis of the object bounding box. The color descriptor is given by the hue histogram of pixels inside the segmented region (16 bins). These descriptors were clustered separately for each object feature to obtain a symbolical description of the object. Clustering was done using a variation of K-means, X-means, that computes the number of clusters from the data \cite{xmeans00pelleg}. Changes in the image were recorded as velocities for the hand and the object. A liner regression was fitted to the trajectories of each experiment and the corresponding coefficients were clustered. Table \ref{tab:clusters} summarizes the obtained clusters.

\begin{table}[!h]
\caption{Summary of symbolic variables and values obtained from clustering.}
\label{tab:clusters}
\centering
\begin{tabular}{llp{3cm}}
\hline\hline
Name 	& Description 	& Values\\ \hline
Action	& Action		& {\em grasp, tap, touch}\\
Color		& Object color	& {\em lightgreen, darkgreen, yellow, blue}\\
Shape	& Object shape		& {\em sphere, box}\\
Size	& Object size		& {\em small, medium, big}\\
ObjVel	   & Object velocity	& {\em slow, medium, fast}\\
HandVel    & Hand velocity  & {\em slow, fast}\\
ObjHandVel & Object-hand velocity					& {\em slow, medium, fast}\\
Contact	   & Contact duration	& {\em short (none), long}\\
\hline\hline
\end{tabular}
\end{table}

\begin{figure}[!t]
 \centering
\includegraphics[width=0.65\columnwidth]{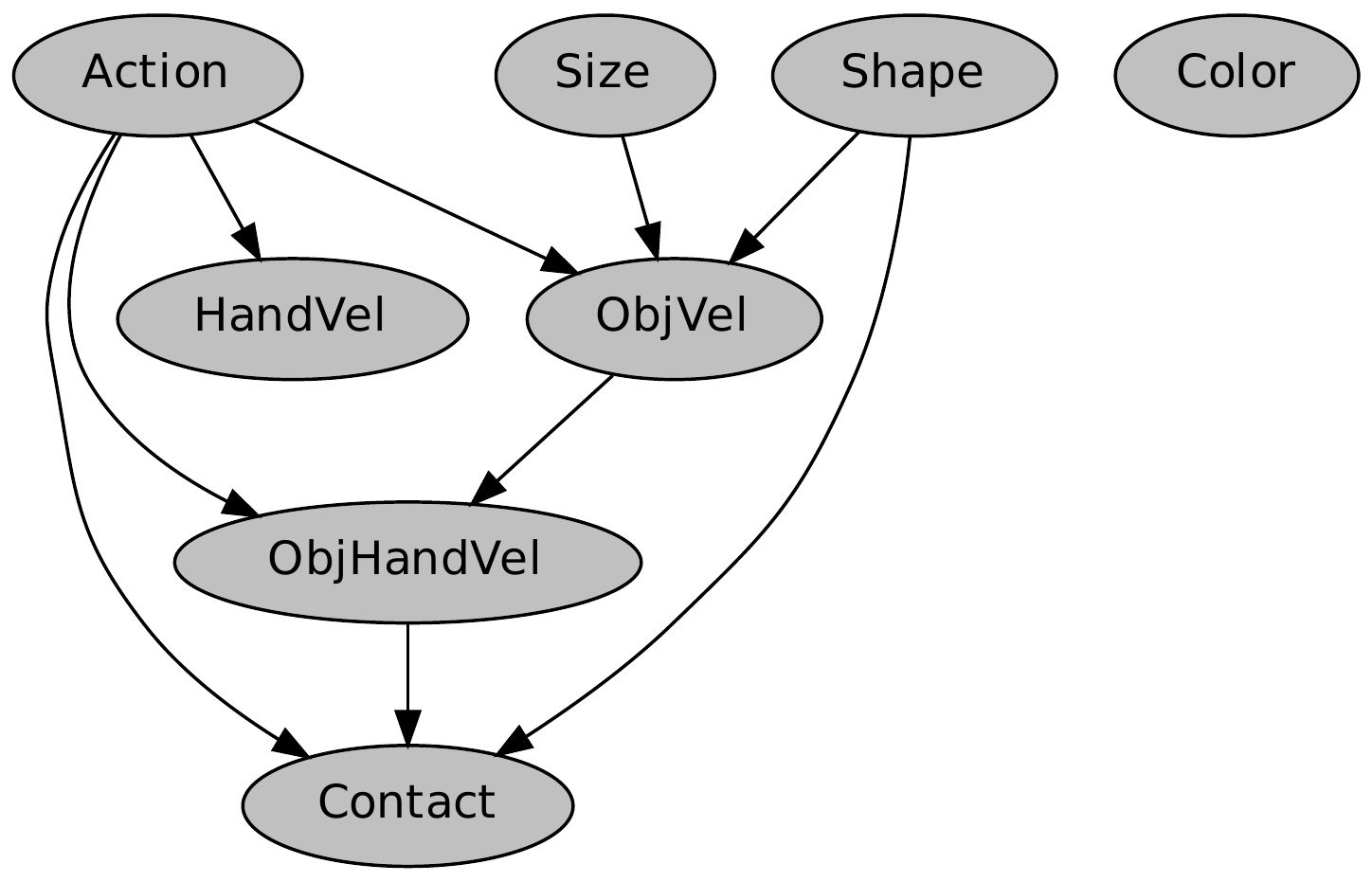}
\caption{Affordances learned by experience.}
\label{fig:affnet}
\end{figure}

Based on the clustered data, the resulting affordance network captures the dependencies between actions, object properties and effects. 
Figure \ref{fig:affnet} shows the affordance network used to discover word meanings.  
Some characteristics are as expected. Color is not linked to any other node since it is irrelevant to the behavior of the object. Shape and size provide for every possible action the corresponding conditional probabilities of effects. However, the links to the effects reflect the specificities of the robot interaction with the objects and, given the used features, the dependencies among them. 
We refer the reader to \cite{Montesano08} for a full description of the experiments and the resulting affordance network. 
It is worth to mention that words can be attached to any network configuration, not only to a specific node or label and, consequently, the affordance network provides the set of possible meanings for the words. 

\subsection{Speech Data} \label{sec:data}
The data from each experiment was augmented with a verbal description describing first the action the robot performs on a certain object and then the effects that the action has produced. Examples of this are: ``Baltazar is grasping the ball but the ball is still.'', ``The robot touches the yellow box and the box is moving.'', ``He taps the green square and the square is sliding.''. Each action, object property and effect is represented by a varying number of synonyms for a total of 49 words. The descriptions were generated automatically from the affordance data using a pseudo-random controlled process in order to randomly pick different synonyms for each concept but retaining a balanced distributions of the words. This procedure was improved comparing to \cite{KrunicEtAl2009ICRA}. Although this procedure is based on a strong simplification of the speech-based interaction, it generates utterances that are complex enough to study the phenomena of interest in the scope of this experiment.

\begin{figure}
 \centering
\includegraphics[width=\columnwidth]{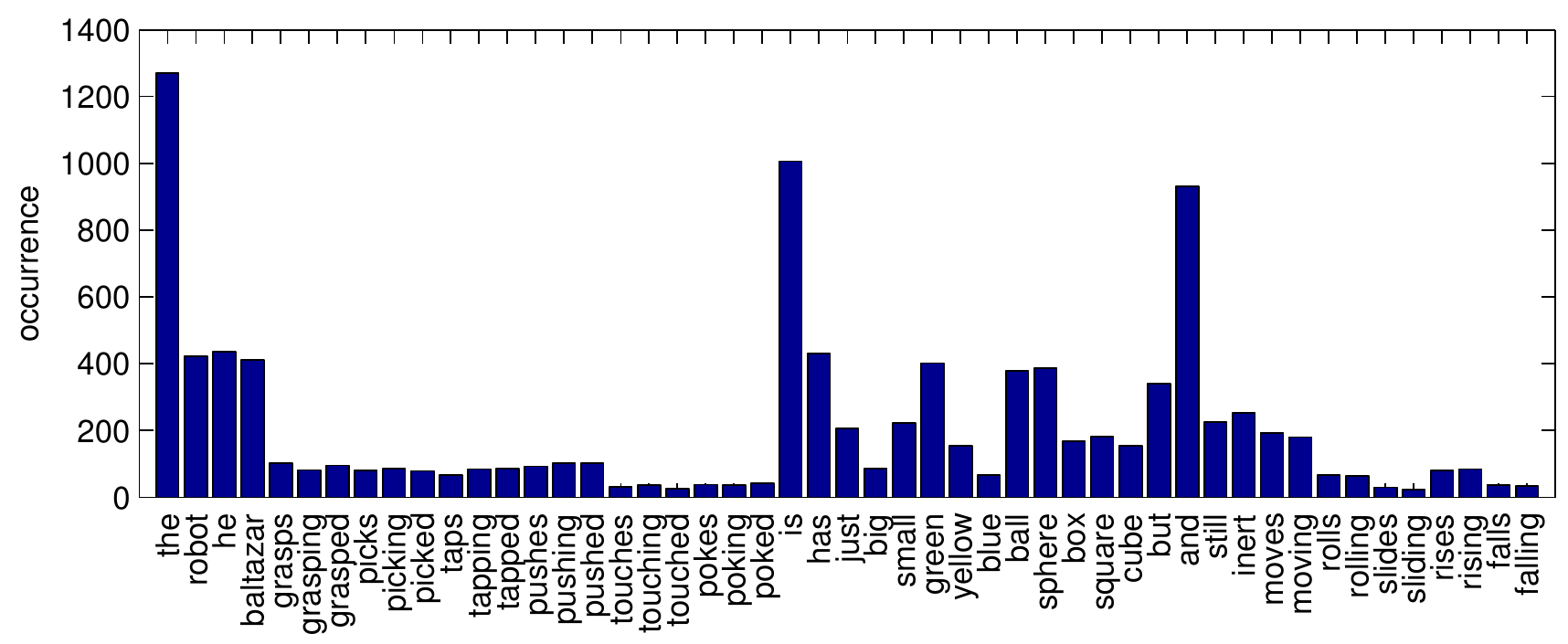}
\caption{Occurrence of words in the $D_w$ dataset (1270 descriptions)}
\label{fig:wordhist}
\end{figure}

The speech recording from \cite{KrunicEtAl2009ICRA} were also disregarded because of technical problems. New recording were performed in a quiet room with a Shure WH20 Dynamic Headset Microphone connected to a TASCAM US-122 sound card. Two speakers, one male and one female, recorded five alternative descriptions for each of the 254 manipulation experiments, for a total of 1270 recording. Figure~\ref{fig:wordhist} shows the distribution of the words in the speech material. The histogram does not count repetitions of the same word in each sentence. This in agreement with the bag-of-word assumption used in the Bayesian network that only observes if a word is present or not in a particular sentence.

Besides the training data described above, a number of 54 sentences were designed in order to test the ability of our model to interpret verbal instructions to the robot. The same 49 words used for the description utterances are used in the set of instructions. Particular care has been put in the design of these sentences to test different aspects of the model. Most instructions are incomplete and ambiguous, only specifying certain aspects of the desired object and action. Some instructions specify an impossible combination of object properties, actions or effects. Other specify only the effect we want to achieve and the model needs to infer the proper action for the proper object. Some of such examples are:
\begin{itemize}
\item ``Tap the green small cube'' (complete instruction where the action and all object features are clearly defined),
\item ``Rise the ball'' (incomplete instruction, ball can be of any color or size, moreover, the robot needs to infer which is the best action to make an object rise),
\item ``Roll the small cube'' (impossible request: cubes can not be made to roll).
\end{itemize}

Because the instructions are often incomplete and ambiguous, it was not straightforward to define what the correct answer from the model should be for evaluation. We asked, therefore, 5 human subjects to give their opinion on what they would expect a robot should do when presented with each of the sentences from the set. The answers indicated the possible set of object properties and actions that were compatible with the given instruction. The subjects mostly agreed on the answers, however, to solve the few cases of disagreement, a majority vote was considered in order to define the reference ground truth.

\begin{table}
\centering
\caption{Example of recognition errors measured in accuracy and bag-of-words}
\label{tab:recexample}
{\setlength{\tabcolsep}{2pt}
\begin{tabular}{l}
  \hline\hline
  Accuracy scoring: (3 substitutions, 1 insertion) \\
  \hline
  {\setlength{\tabcolsep}{2pt}\scriptsize
    \begin{tabular}{lllllllllllllll}
      LAB: & the & \textbf{robot} & is & grasping & the & big &                & \textbf{yellow} & sphere & but & \textbf{the} & sphere & is & inert \\
      REC: & the & \textbf{but}   & is & grasping & the & big & \textbf{still} & \textbf{is}     & sphere & but & \textbf{is}  & sphere & is & inert 
    \end{tabular}
  } \\ \hline
  Bag-of-words scoring: (2 false rejections and 1 false acceptance)\\
  \hline
  {\setlength{\tabcolsep}{2pt}\scriptsize
    \begin{tabular}{lllllllllll}
      LAB: & big & but & grasping & inert & is & \textbf{robot} & sphere &                & the & \textbf{yellow} \\
      REC: & big & but & grasping & inert & is &                & sphere & \textbf{still} & the &           
    \end{tabular}
  } \\
  \hline\hline
\end{tabular}
}
\end{table}

\begin{figure}
\centering
\includegraphics[width=0.94\textheight, angle=90]{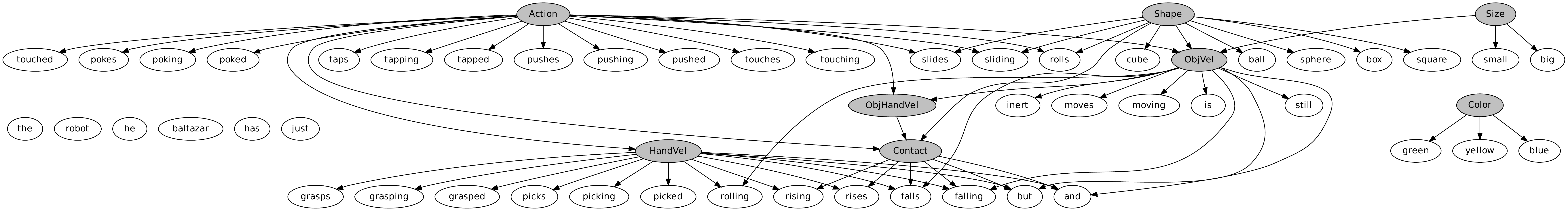}\hspace{2em}
\includegraphics[width=0.94\textheight, angle=90]{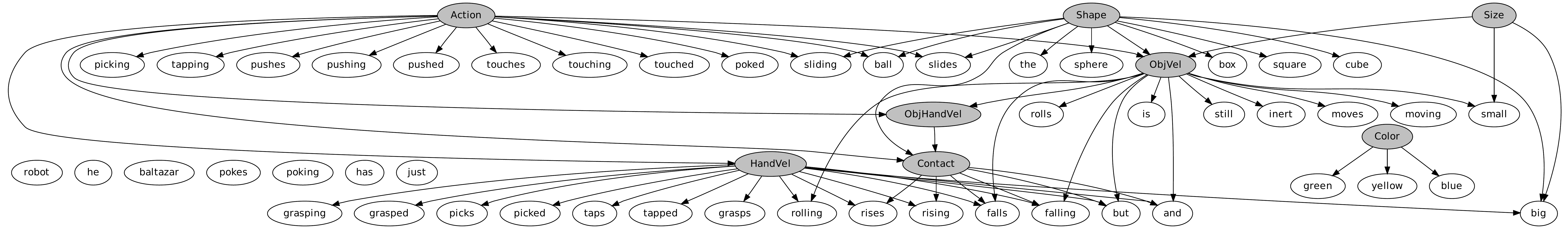}
\caption{Graph of the full Bayesian network. Left: network obtained with labeled speech data. Right: network obtained with recognized speech data.}
\label{fig:fullnets}
\end{figure}

\begin{figure*}
\centering
\includegraphics[width=0.45\columnwidth]{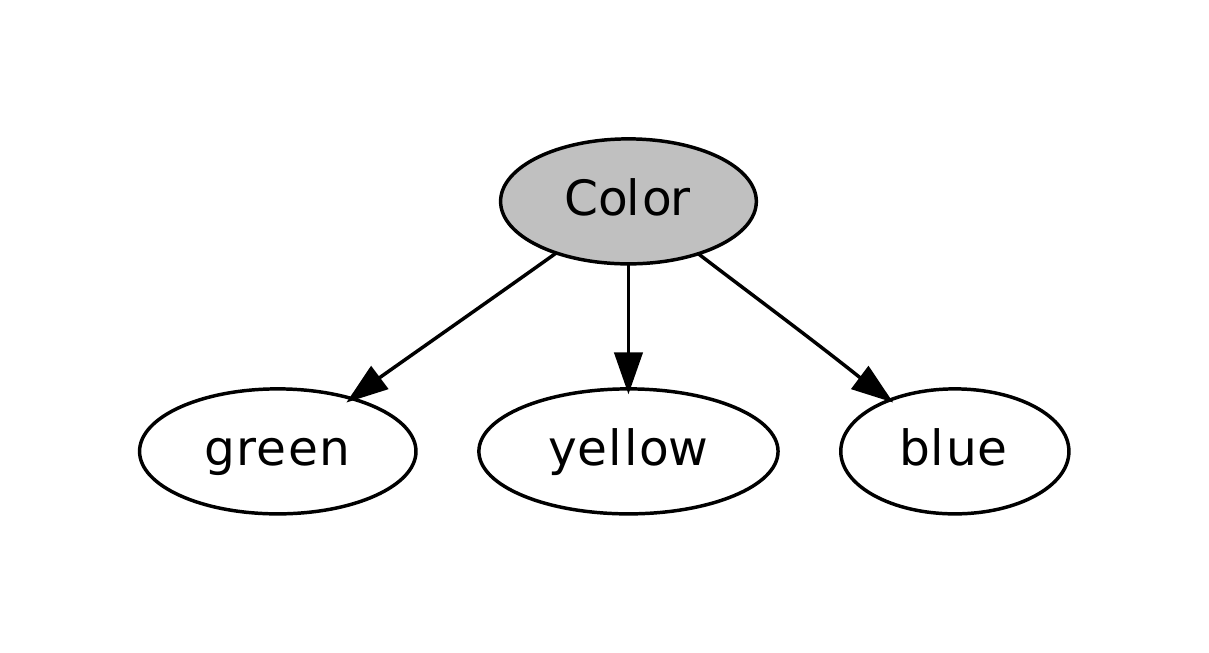}\hspace{1cm}
\includegraphics[width=0.28\columnwidth]{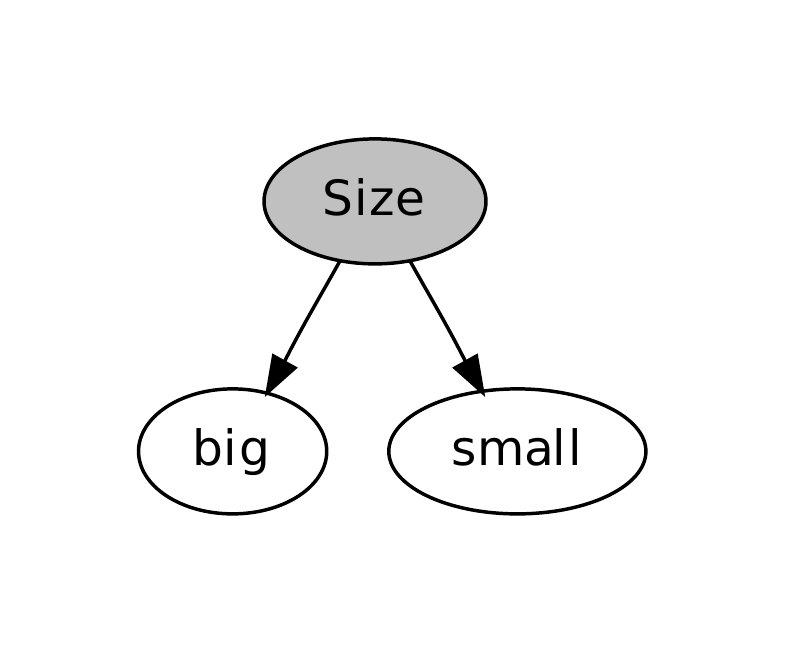}
\includegraphics[width=0.7\columnwidth]{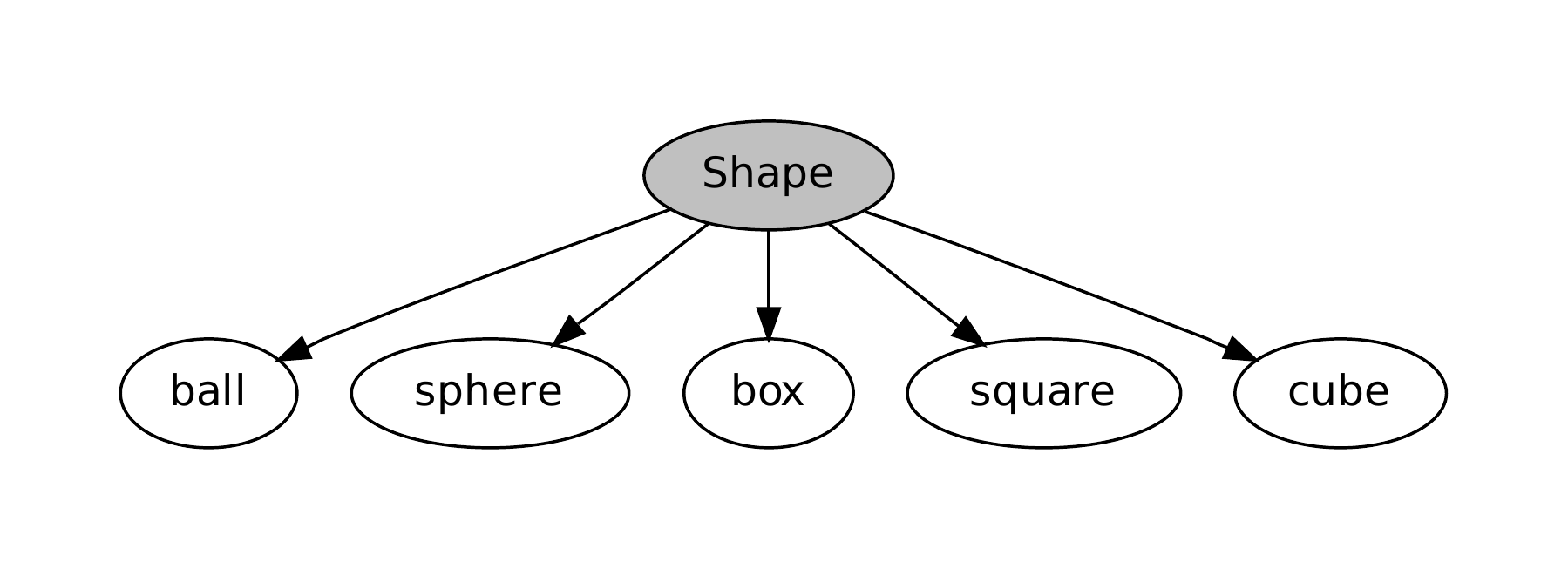}\\
\includegraphics[width=0.36\columnwidth]{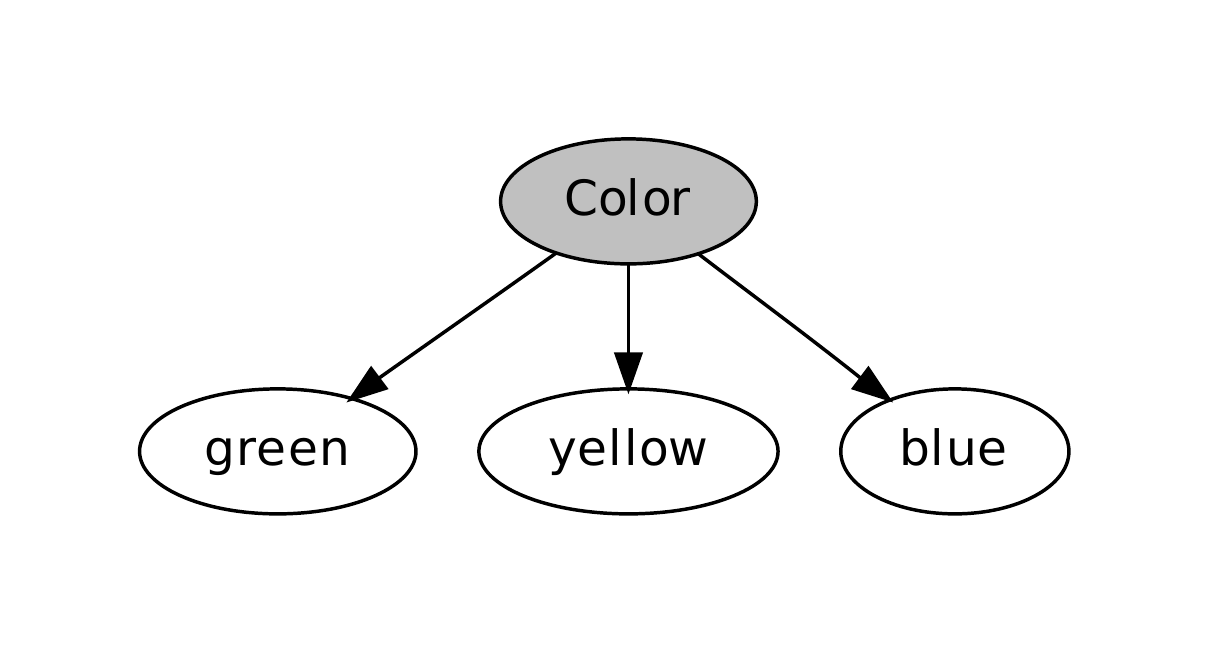}
\includegraphics[width=0.2\columnwidth]{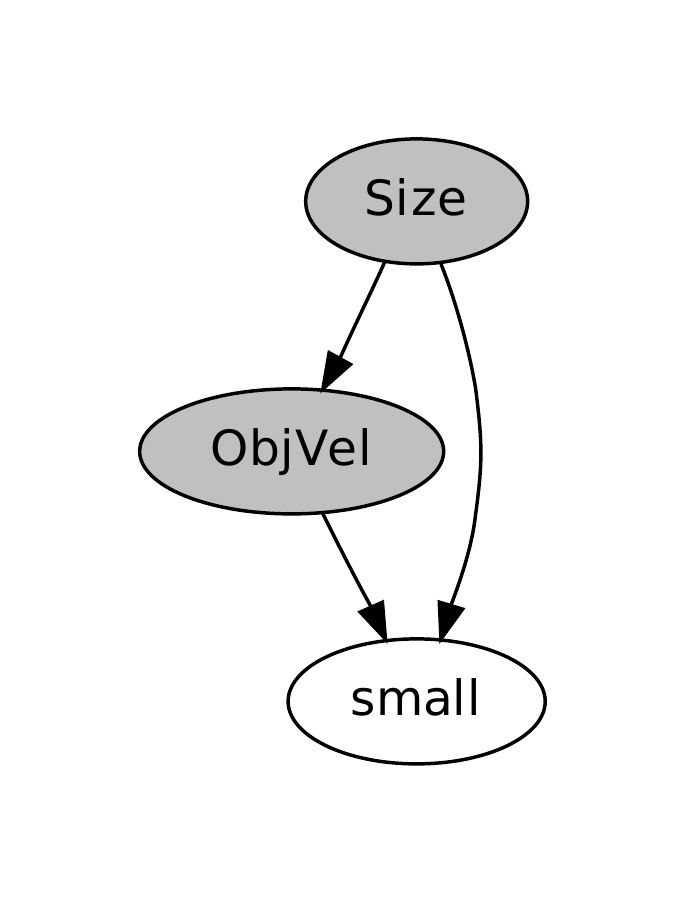}
\includegraphics[width=0.4\columnwidth]{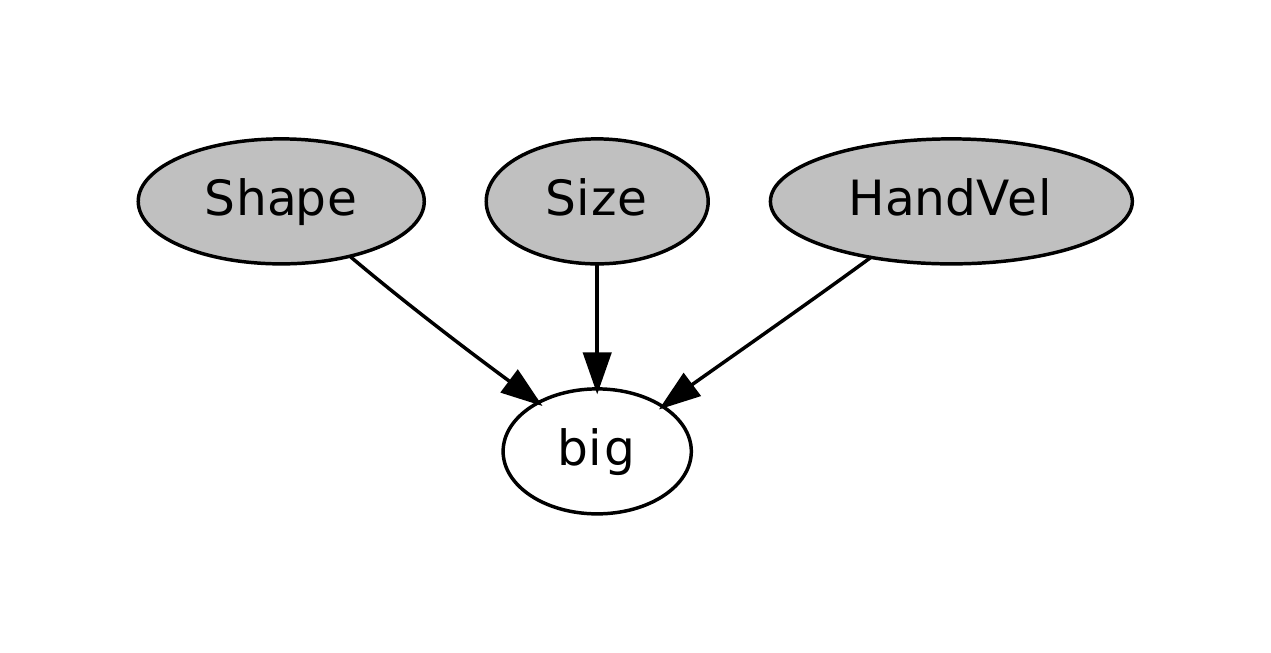}
\includegraphics[width=0.7\columnwidth]{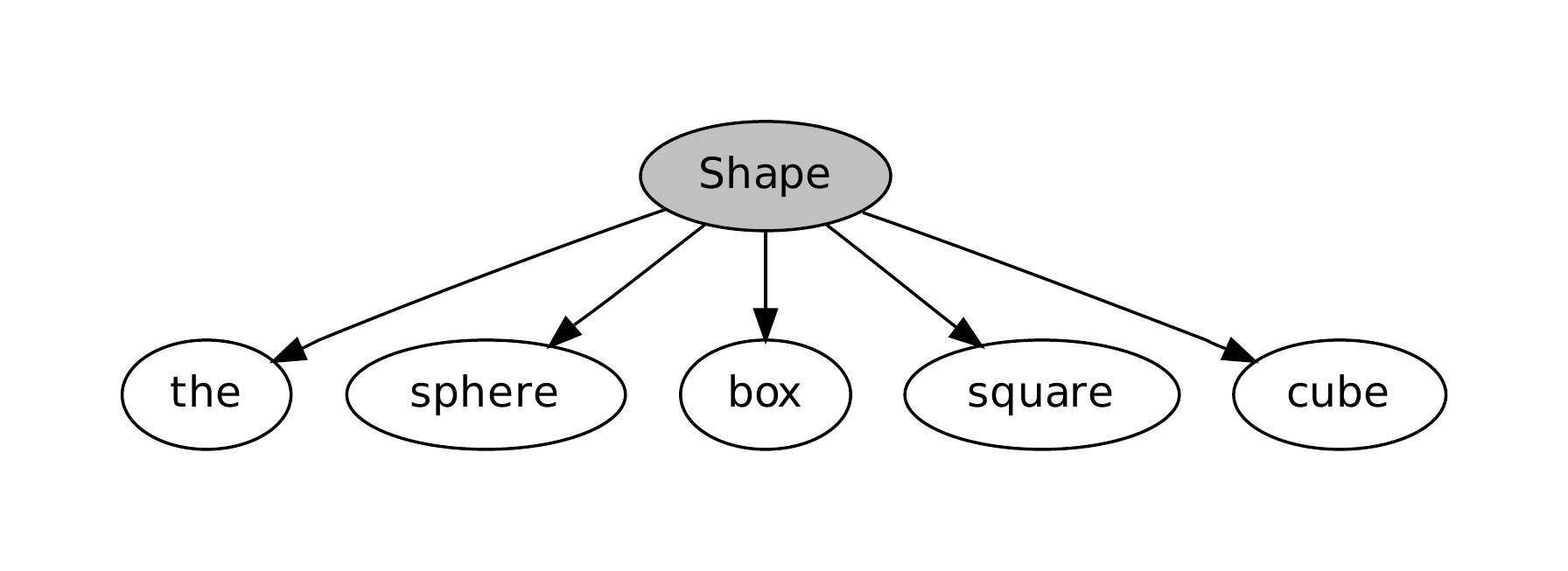}
\caption{Object properties words, top: labeled speech data, bottom: recognized speech data}
\label{fig:objectfeatures}
\vspace{-5mm}
\end{figure*}

\begin{figure*}
\centering
\includegraphics[width=.9\textwidth]{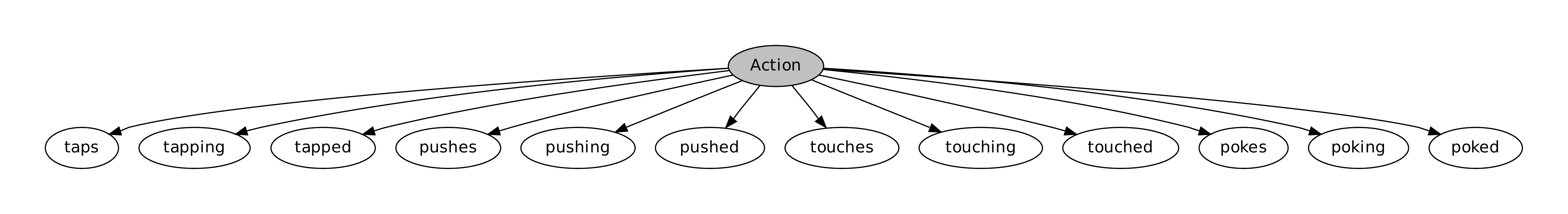}
\includegraphics[width=.7\textwidth]{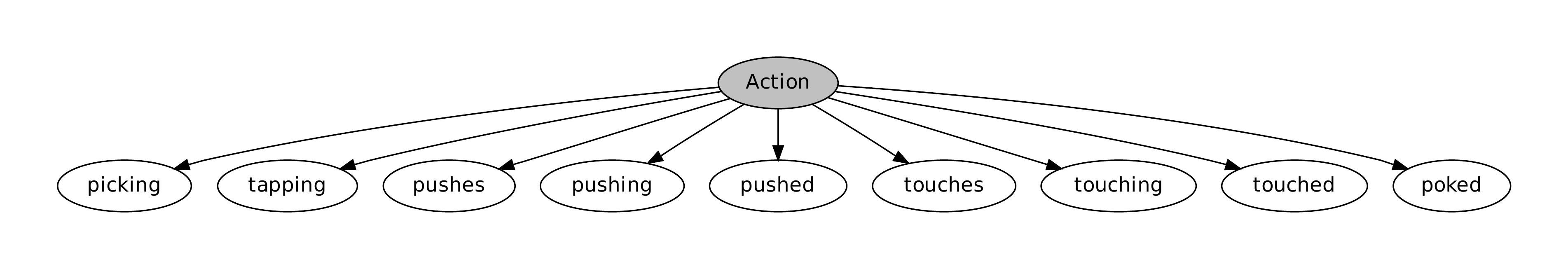}
%\vspace*{-1cm}
\caption{Action words (excluding grasping), top: labeled speech data, bottom: recognized speech data}
\label{fig:actionnograsp}
\vspace{-5mm}
\end{figure*}

\subsection{Speech input}
\label{sec:speechinput}
As discussed in Section~\ref{sec:introduction}, we assume that one of the basic skills of the robot is the ability to classify speech input into sequences of words.

The speech-to-text unit is implemented as a hidden Markov model (HMM) automatic speech recognizer (ASR). Each word belonging to the language described above is modeled as a sequence of phonemes each modeled by a left-to-right HMM. Additionally a three-state-model is defined in order to model silence. Speaker independent models from \cite{gs:Lindberg2000} are used, but the recognizer adapts automatically to new voices by means of Maximum Likelihood Linear Regression (MLLR) adaptation \cite{LeggetterAndWoodland1995}. This adaptation scheme is unsupervised in the sense that does not make use of the knowledge about what was said in the sentences. For this reason it is compatible with our developmental approach.

During recognition, no grammatical structure other than a simple loop of words was imposed to the decoder at run time, in agreement with our hypothesis that a grammar is not necessary in order to learn simple word-meaning associations. Furthermore, the sequence of words output by the recognizer is ``flattened'' in order to be input to the Bayesian network. This means that out of each sentence, a Boolean vector is constructed solely indicating if the word was or was not present in the sentence.

The performance of the recognizer was computed in two different ways as illustrated in Table~\ref{tab:recexample}. The first is standard in ASR research and is similar to the Levenshtein distance. It is achieved by realigning the reference and recognized sentences with dynamic programming, as illustrated in the upper part of Table~\ref{tab:recexample}, and counting the number of insertions $I$, deletions $D$, and substitutions $S$ (bold in the table). A global accuracy score is then computed as $ A = \frac{N-D-S-I}{N} = \frac{H-I}{N} $ where $N$ is the total number of reference words and $H$ the number of correct words. The second, more tightly connected to our task, is a simple classification rate in the bag-of-words assumption. In this case we count for each utterance the number of false rejections and false acceptances over the number of unique words in that utterance. The accuracy of the recognizer was 83.9\% and the bag-of-words classification rate is 82.7\% (with 8.9\% false rejections and 8.3\% false acceptance). If we compute the bag-of-word classification rate over the size of the vocabulary instead of the utterance lengths, we obtain 96.8\% (with 1.7\% false rejections and 1.5\% false acceptance). Perhaps a better indication of the amount of errors from the recognizer is that, in average, there is a false acceptance every 1.3 utterances and a false rejection every 1.2 utterances.

% do not remove (emacs configuration)
% Local variables:
% enable-local-variables: t
% ispell-local-dictionary: "british"
% mode: latex
% eval: (flyspell-mode)
% eval: (flyspell-buffer)
% End:

\section{Results}
\label{sec:results}

This section presents different aspects of the results obtained in our experiments. Firstly, we analyze the structure learned by the Bayesian network as an indication of word-meaning associations acquired by the robot. Secondly, we analyze the use of the model in practical applications, such as interpreting instructions or using context in order to improve speech recognition. In all cases we compared the model learned on the transcribed data, also called labeled speech data, to the one learned from the automatically recognized speech data.

\subsection{Learning}
The results of learning word meaning associations are displayed in Figure~\ref{fig:fullnets} and detailed in the following figures. Figure~\ref{fig:fullnets} displays the full graph of the Bayesian network, where the affordance nodes are filled whereas word nodes have white background. Both the network learned from labeled data and from recognized data are shown. As explained in Section~\ref{sec:speechinput}, the difference between labeled and recognized data is that the recognizer may either miss certain words or insert extra words in each linguistic description. The full networks are included to give an impression of the overall complexity of the model. In the following, we will focus on subsets of words, by only displaying parts of the networks in Figure~\ref{fig:fullnets}, in order to simplify the discussion.

Some of the word nodes do not display any relationship with the affordance nodes. The so called \emph{non-referential} words are: ``robot'', ``just'', ``the'', ``he'', ``Baltazar'', ``has''. This result is not surprising if we notice that the affordance network did not include a representation of the robot itself (``robot'', ``he'', ``Baltazar''), nor a representation of time (``just''). Moreover, articles and auxiliary verbs were also expected to be non-referential. When ASR data is used for learning, in addition to the above non-referential words, the words ``pokes'' and ``poking'' also appear to have no connection with the affordance nodes. In the labeled data, ``pokes'' and ``poking'' appear 74 times consistently in connection to the action touch. However, these words are most often misrecognized by the recognizer and in the ASR data they appear only 11 times of which 6 times in connection with the action touch, 4 times with grasp and once with tap.

\begin{figure}
\centering
\includegraphics[width=0.8\columnwidth]{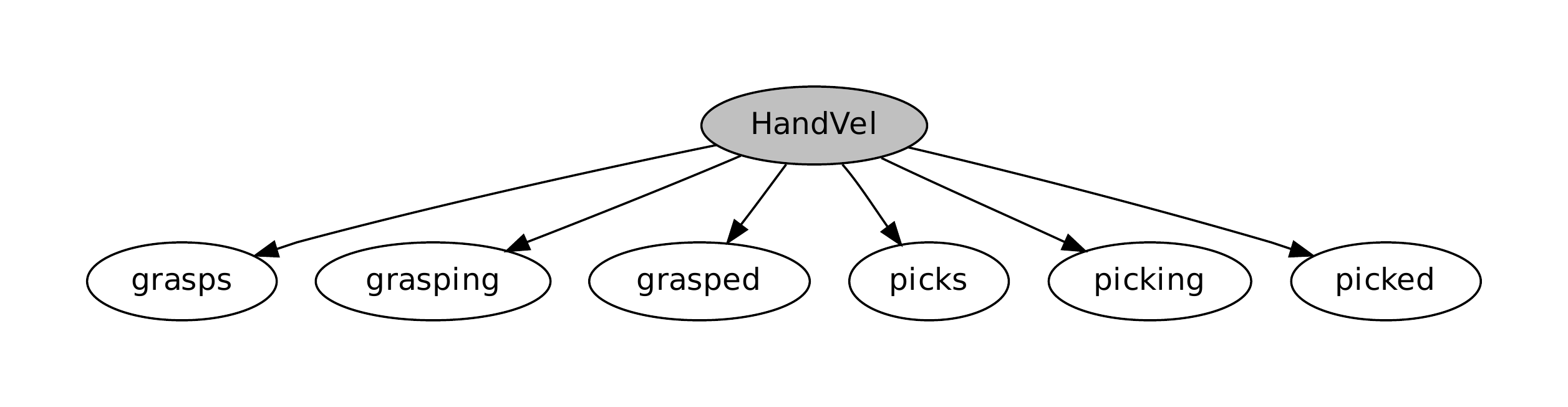}
\includegraphics[width=0.98\columnwidth]{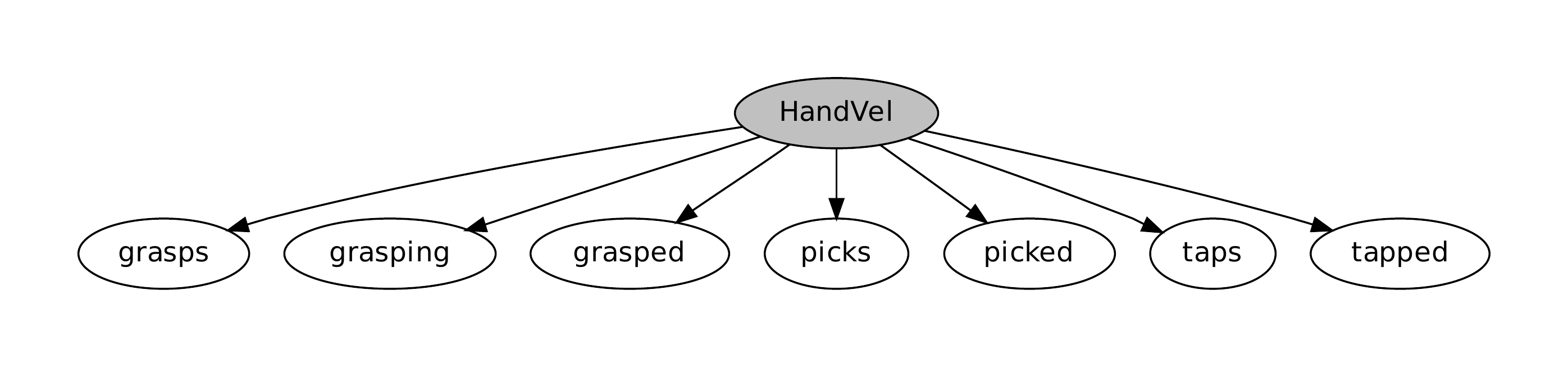}
\vspace*{-5mm}
\caption{Action words (grasping)}
\label{fig:actiongrasp}
\end{figure}
\begin{figure}
\centering
\includegraphics[width=0.7\columnwidth]{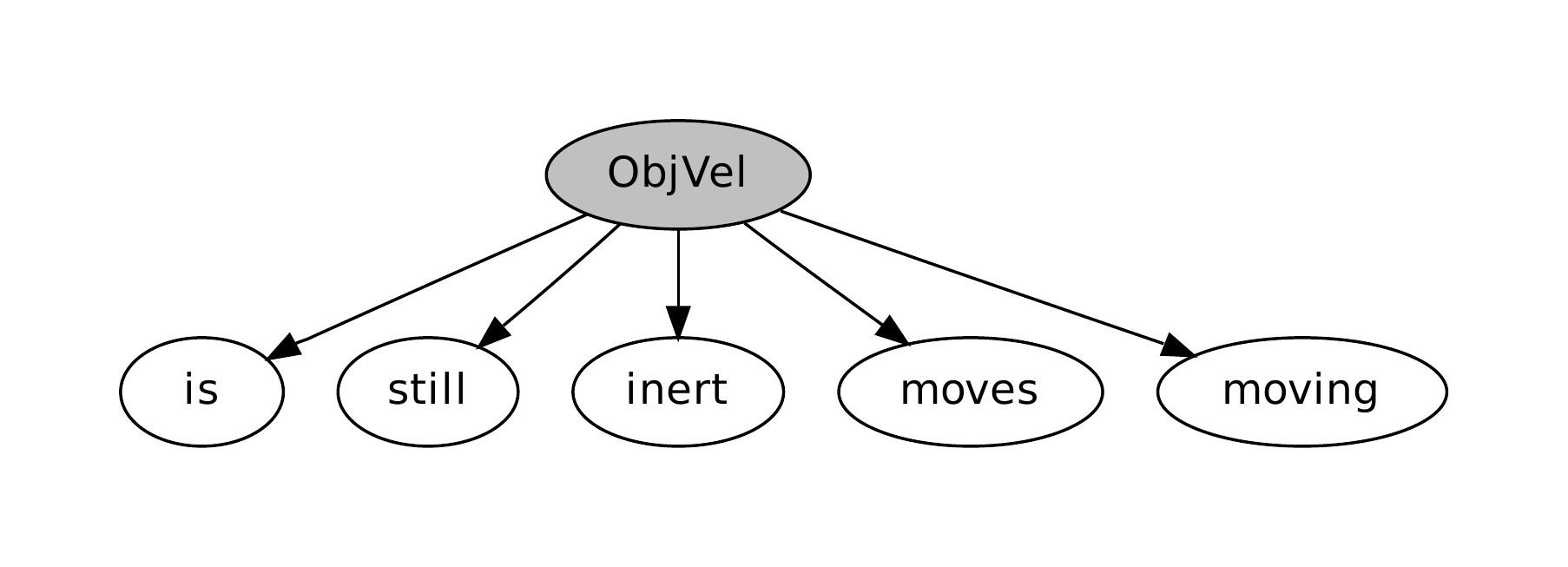}
\includegraphics[width=0.8\columnwidth]{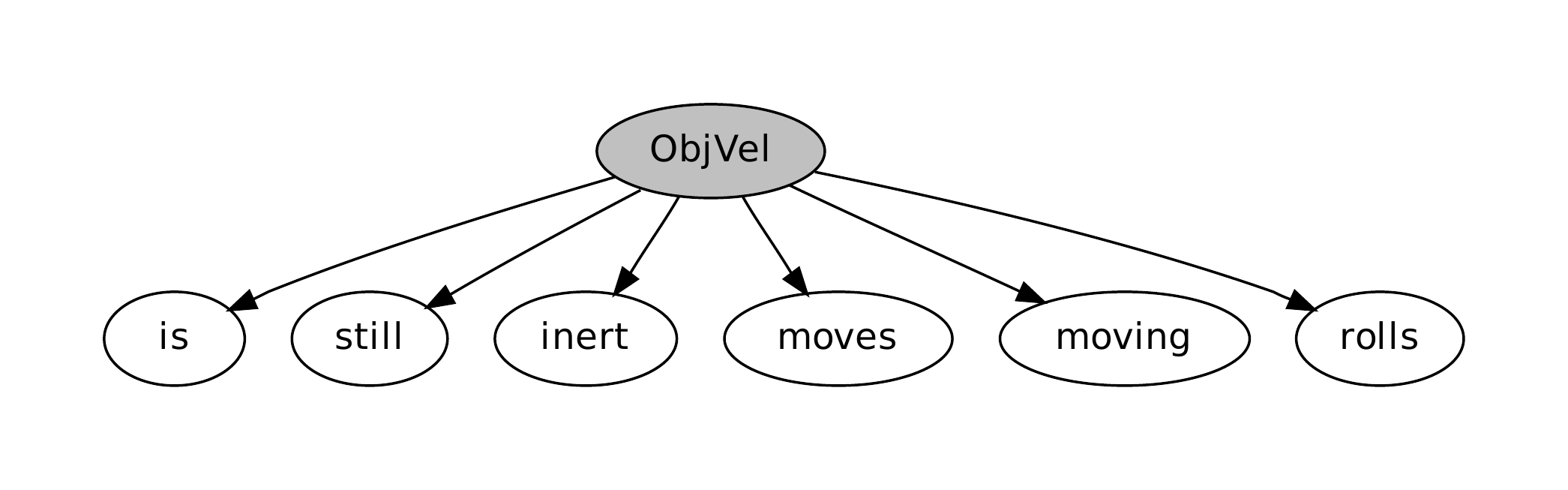}
\vspace*{-5mm}
\caption{Effect words: generic movement, top: labeled speech data, bottom: recognized speech data}
\label{fig:genericmovement}
\end{figure}

Words expressing \emph{object features} are displayed in Figure~\ref{fig:objectfeatures} (top) for learning from labeled data. These are clearly linked to the right affordance node. This result is in accordance with previous research that showed that it is possible to learn word object associations. However, the structure is not as clean for the ASR data, as we can see in Figure~\ref{fig:objectfeatures} (bottom). In particular, the size-related words (``small'', ``big'') are not only connected to the Size node, but to spurious nodes such as ObjVel, Shape and HandVel.

Top of Figure~\ref{fig:actionnograsp} shows the words that were linked to the Action node in the labeled data learning. These include all the action words apart from the words referring to the action grasp that are treated differently by the model (see later). The ASR case, shown in the bottom plot, is the same apart from the words ``pokes'' and ``poking'' discussed above, and the words ``touching'', ``taps'' and ``tapped''.

Words corresponding to the action grasp are linked by the model to the node Hand Velocity (HandVel) as shown in Figure~\ref{fig:actiongrasp} for both labeled data and recognized data. The reason for this is that, in our data, HandVel is high only for grasping actions. The information on hand velocity is, therefore, sufficient to determine whether a grasp was performed. Moreover, HandVel can only assume two values (high and low, as a result of the sensory input clustering), while Action can assume three values (grasp, tap and touch), thus making the first a more concise representation of the concept grasp. In the ASR case also the words ``taps'' and ``tapped'' are connected to this node probably due to recognition errors.

Words describing \emph{effects} usually involve more affordance nodes. In case of words indicating generic movement the link is to the object velocity node, as expected (see Figure~\ref{fig:genericmovement}). Note also that the auxiliary verb ``is'' is connected to this node because it is only used in the expressions of movement such as ``is still'' or ``is moving'' in our data.

\begin{figure}
\centering
\includegraphics[width=0.40\columnwidth]{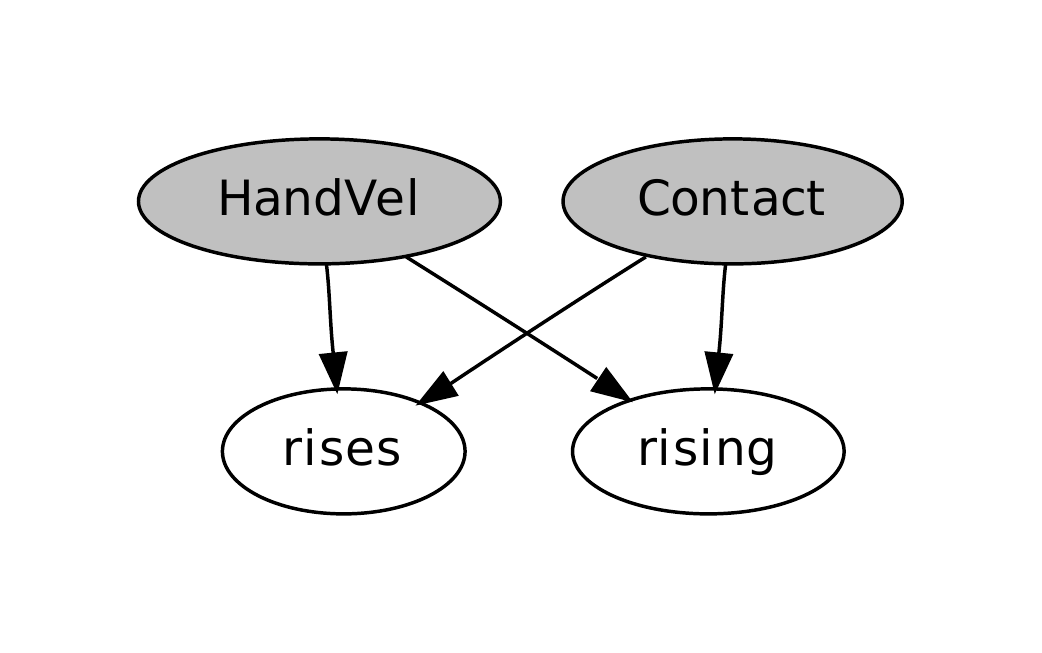}
\includegraphics[width=0.56\columnwidth]{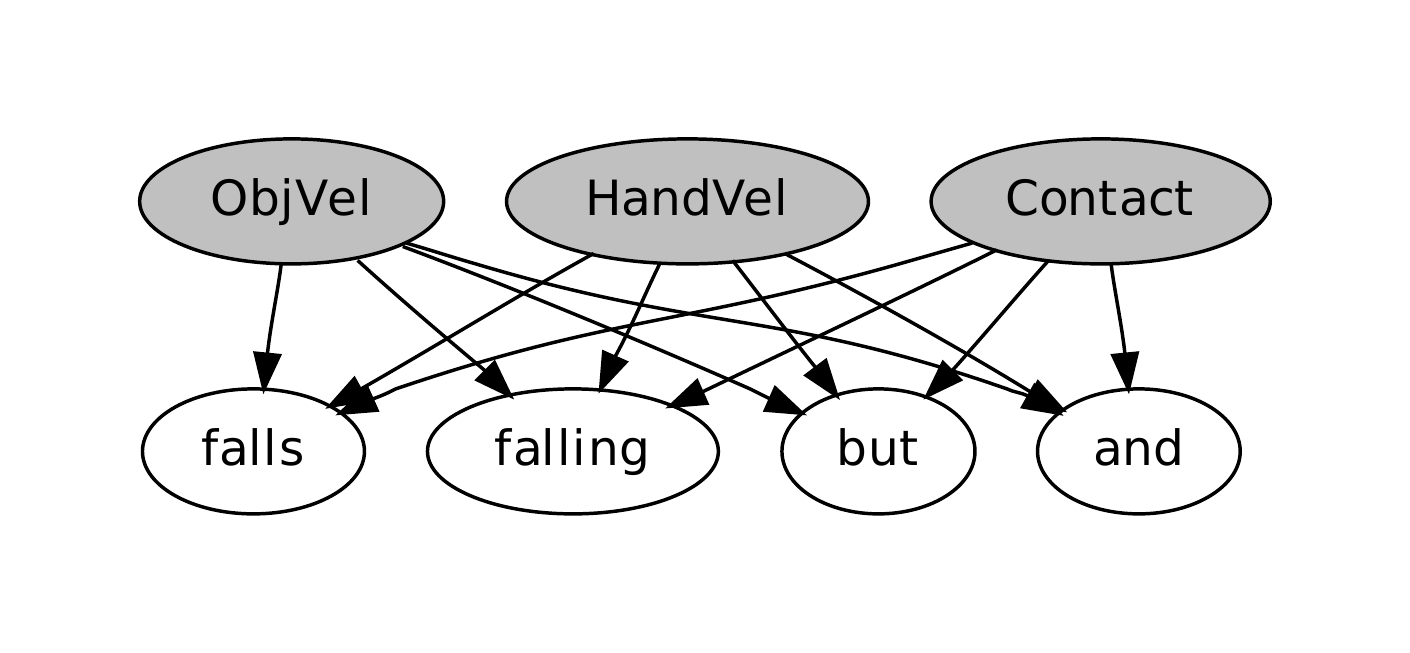}
\caption{Effect words: vertical movement, both for labeled and recognized data}
\label{fig:verticalmovement}
\end{figure}

Words describing vertical movement are shown in Figure~\ref{fig:verticalmovement}. Interestingly in this case, exactly the same association is obtained with the labeled speech data and with the recognized speech data. In order to understand these associations we have to consider that vertical movement in our data is only obtained in case of attempted grasp. This is why Hand Velocity and hand-object Contact are involved. The reason why in case of falling objects we also need the Object Velocity is probably that a short Contact is not enough to specify if the grasp failed from the beginning and, therefore, the object is inert, or if it failed after having lifted the object, thus making it fall. Also, the reason why ``and'' and ``but'' are treated like ``falls'' and ``falling'' is that the conjunction is chosen depending on the success of the action, and the action grasp is the most likely to fail in our experiments.

\begin{figure}
\centering
\includegraphics[width=0.52\columnwidth]{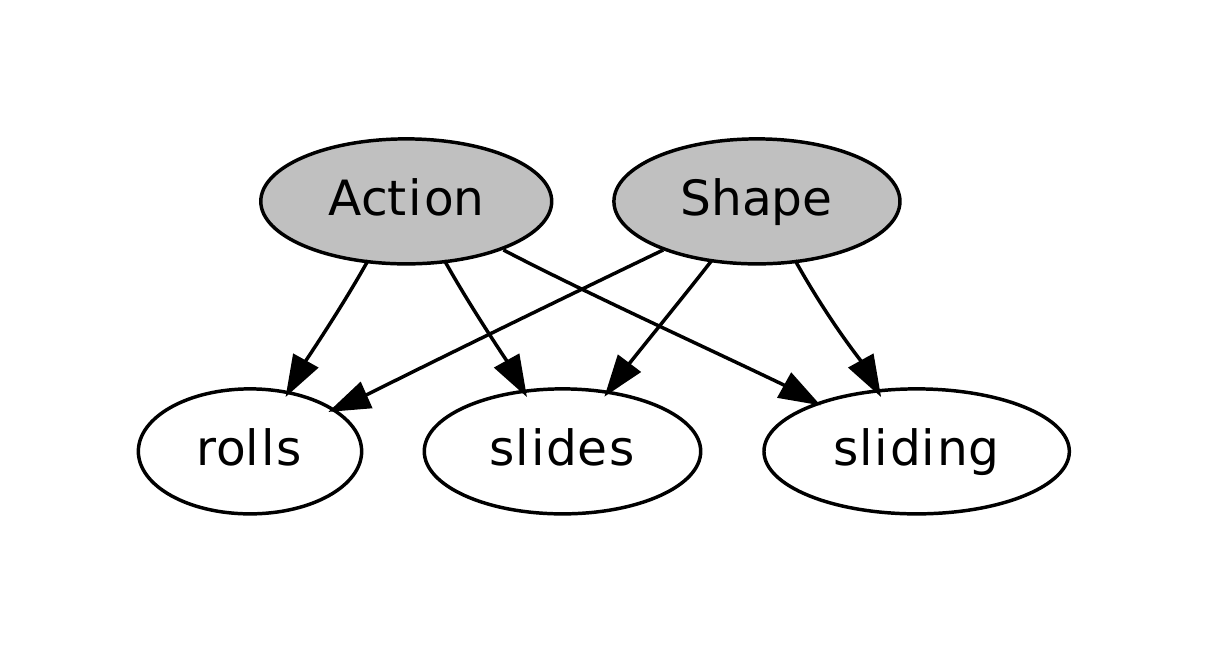}
\includegraphics[width=0.44\columnwidth]{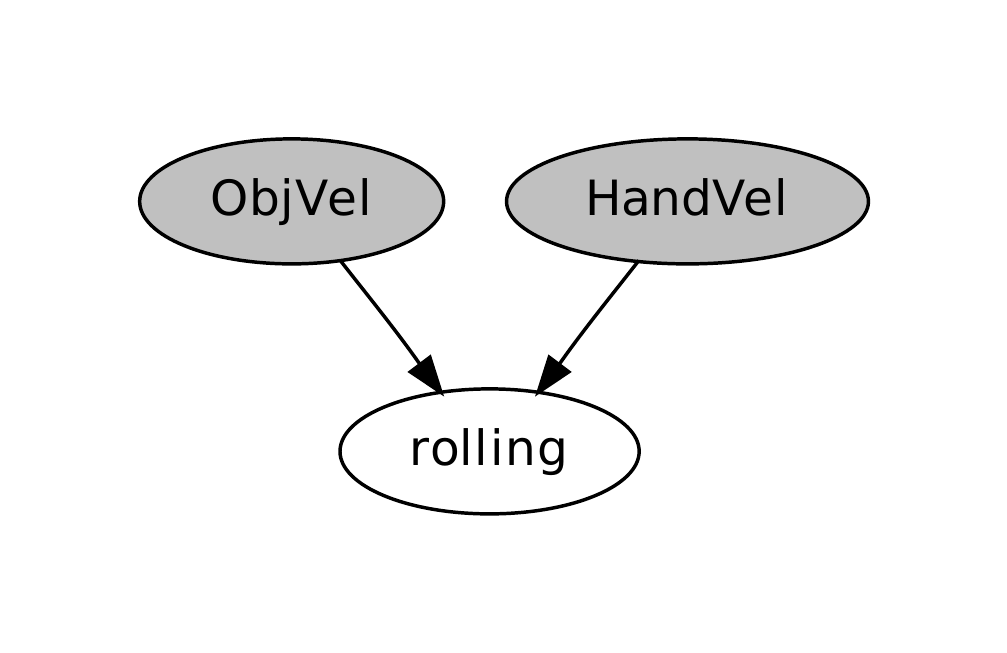}
\includegraphics[width=0.52\columnwidth]{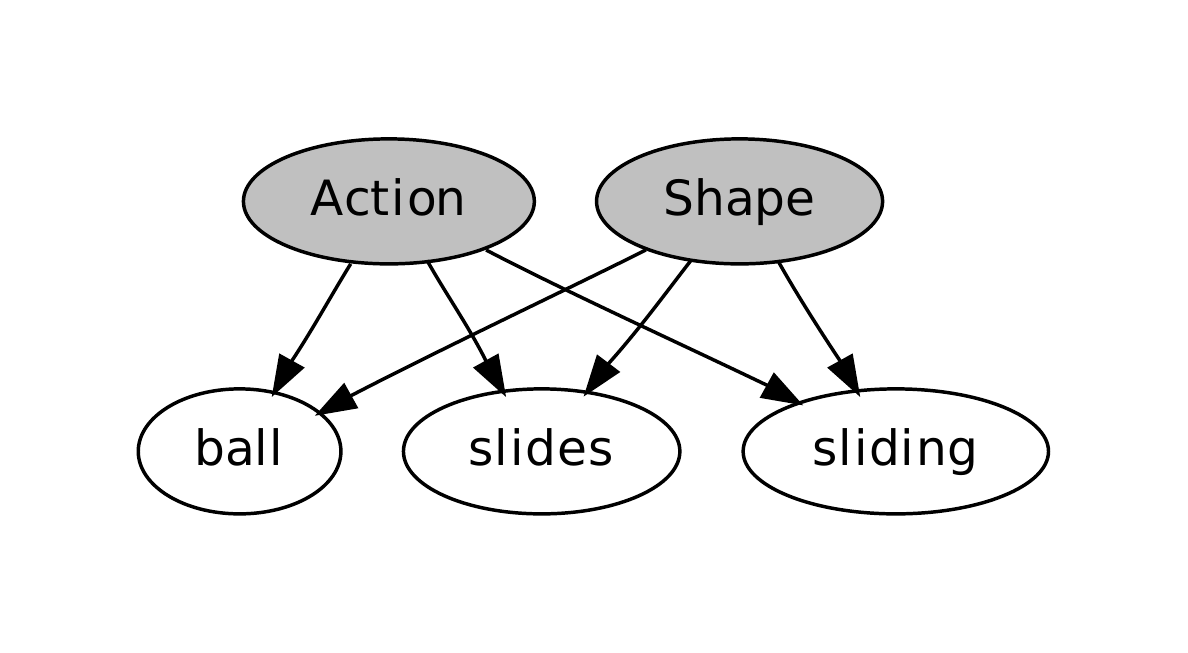}
\includegraphics[width=0.44\columnwidth]{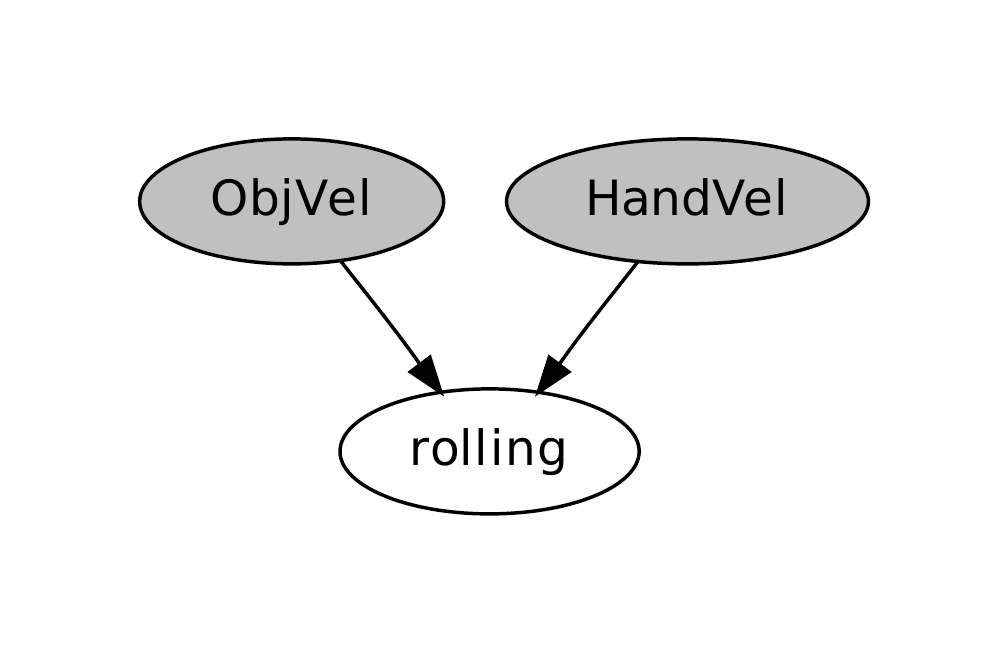}
\caption{Effect words: horizontal movement, top: labeled speech data, bottom: for recognized speech data}
\label{fig:horizontalmovement}
\end{figure}

Finally the horizontal movements are displayed in Figure~\ref{fig:horizontalmovement}. These are also treated similarly in the labeled data and recognized data case. The only difference is that the words ``rolls'' and ``ball'' are switched.

\begin{table*}
\caption{Examples of using the Bayesian network to select actions and objects}
\label{tab:useexamples}
\centering
(a) Network trained with labeled speech data\par
\begin{tabular}{l|ccccccc}\hline
 objects on the table & \multicolumn{7}{c}{Verbal input}\\
 (cluster lables) & ``small grasped''& ``moving green''& ``ball sliding''& ``big rolling''& ``has rising''& ``sliding small''& ``rises yellow'' \\\hline
 lightgreen big sphere & - & grasp, p=0.01 & - & \textbf{tap, p=0.12} & grasp, p=0.01 & - & - \\
 yellow medium sphere & - & - & - & - & grasp, p=0.10 & - & \textbf{grasp, p=0.40} \\
 darkgreen small box & grasp, p=0.18 & grasp, p=0.04 & - & - & grasp, p=0.05 & \textbf{tap, p=0.48} & - \\
 blue medium box & - & - & - & - & grasp, p=0.01 & - & - \\
 blue big box & - & - & - & touch, p=0.01 & - & - & - \\
 darkgreen small sphere & \textbf{grasp, p=0.30} & \textbf{tap, p=0.10} & - & - & \textbf{grasp, p=0.12} & - & - \\
\hline
\end{tabular}
\vspace{3mm}

(b) Network trained with recognized speech data\par
\begin{tabular}{l|ccccccc}\hline
 objects on the table & \multicolumn{7}{c}{Verbal input}\\
 (cluster lables) & ``small grasped''& ``moving green''& ``ball sliding''& ``big rolling''& ``has rising''& ``sliding small''& ``rises yellow'' \\\hline
 lightgreen big sphere & - & grasp, p=0.01 & - & tap, p=0.05 & grasp, p=0.01 & - & - \\
 yellow medium sphere & grasp, p=0.05 & grasp, p=0.01 & - & tap, p=0.02 & grasp, p=0.10 & - & \textbf{grasp, p=0.33} \\
 darkgreen small box & grasp, p=0.12 & grasp, p=0.04 & \textbf{tap, p=0.17} & \textbf{tap, p=0.08} & grasp, p=0.05 & \textbf{tap, p=0.38} & - \\
 blue medium box & grasp, p=0.01 & - & tap, p=0.05 & tap, p=0.02 & grasp, p=0.01 & tap, p=0.02 & - \\
 blue big box & - & - & tap, p=0.01 & touch, p=0.01 & - & - & - \\
 darkgreen small sphere & \textbf{grasp, p=0.22} & \textbf{tap, p=0.10} & - & grasp, p=0.01 & \textbf{grasp, p=0.12} & - & - \\
\hline
\end{tabular}
\end{table*}

\begin{table} 
\caption{Examples of using the Bayesian network to improve ASR}
\label{tab:useexamples2}
\centering
(a) Network trained with labeled speech data\par
\begin{tabular}{B|CCC}\hline
 & \multicolumn{3}{c}{N-best list from ASR (N=3)}\\\hline
 objects on the table (cluster lables) & ``tapping small sliding'' \textbf{p=0.100}& ``tapping box slides'' p=0.070& ``tapped ball rolls'' p=0.010 \\\hline
 lightgreen big sphere & 0.0 & 0.0 & 3.409E-03 \\
 yellow medium sphere & 0.0 & 0.0 & 2.926E-03 \\
 darkgreen small box & 1.500E-03 & 1.357E-03 & 0.0 \\
 blue medium box & 0.0 & 1.260E-03 & 0.0 \\
 blue big box & 0.0 & 1.481E-03 & 0.0 \\
 darkgreen small sphere & 0.0 & 0.0 & 2.926E-03 \\
\hline
final score & 1.500E-04 & \textbf{2.868E-04} & 9.261E-05 \\\hline
\end{tabular}
\vspace{3mm}

(b) Network trained with recognized speech data\par
\begin{tabular}{B|CCC}\hline
 & \multicolumn{3}{c}{N-best list from ASR (N=3)}\\\hline
 objects on the table (cluster lables) & ``tapping small sliding'' \textbf{p=0.100}& ``tapping box slides'' p=0.070& ``tapped ball rolls'' p=0.010 \\\hline
 lightgreen big sphere & 0.0 & 0.0 & 2.965E-03 \\
 yellow medium sphere & 0.0 & 0.0 & 5.149E-03 \\
 darkgreen small box & 1.708E-03 & 1.553E-03 & 3.496E-04 \\
 blue medium box & 2.647E-04 & 1.447E-03 & 3.226E-04 \\
 blue big box & 5.366E-05 & 1.691E-03 & 1.418E-04 \\
 darkgreen small sphere & 0.0 & 0.0 & 5.248E-03 \\
\hline
final score & 2.027E-04 & \textbf{3.283E-04} & 1.418E-04 \\\hline
\end{tabular}
\end{table}

Note that, although a clean and interpretable network structure is a desirable property, we should not focus only on the structure. First, some dependencies cannot be directly explained without taking into account the robot capabilities, which differ from that of humans. Second, in a noisy environment, there are likely to be spurious connections as the ones we can see in the figures above. These may or may not be removed with more experience (more observations), however, what is interesting to evaluate is the ability of the model to predict the correct or reasonable answer in spite of the noisy structure, as we will see in the following.

\subsection{Using the model}
As noted in the previous Section, the evaluation of the model should be done considering its use in practical applications. For this reason, we performed some prediction experiments were we test different ways that the model can be used to perform inference.

Table~\ref{tab:useexamples} shows some examples of using incomplete verbal descriptions to assign a task to the robot. Table~\ref{tab:useexamples}~(a) is obtained with a model trained on the labeled speech data, whereas the results in Table~\ref{tab:useexamples}~(b) are obtained with recognized speech data. The robot has a number of objects in its sensory field (represented by the object features in the first column in the Table). The Table shows, for each verbal input $W_S$ (column) and each set of object features $F_{o_i}$ (row), the best action computed by Equation~\ref{eq:selectactionobject} when the set of objects $O_s$  is restricted to a specific object ${o_i}$. The global maximum over all actions and objects for a given verbal input, corresponding to the general form of Equation~\ref{eq:selectactionobject}, is indicated in bold face in the table. Also, to simplify the table, probabilities that are below the two-digit precision shown, are displayed as dashes instead of zeros.

If the combination of object features and verbal input is incompatible with any actions, $P(a_i, F_{o_i}\mid W_S)$ may be 0 $\forall a_i\in\mathcal{A}$. In case this happens for all available objects (as for ``ball sliding'' in Table~\ref{tab:useexamples}~(a)), the behavior of the robot is not defined. A way to solve such cases may be, e.g., to initiate an interaction with the human in order to clarify his/her intentions. Note, however, that the ability of the model to detect these inconsistent inputs is reduced when we used noisy recognized data to train it. In this case the model may have seen inconsistent input due to the recognition errors in the training phase, and, therefore, output nonzero probabilities as in the third column of Table~\ref{tab:useexamples}~(b). Besides these cases, Table~\ref{tab:useexamples} shows that, in spite of the different structure shown in the previous Section, the model generates very similar inferences given the same observations.

Another application of our model is to use the knowledge stored in the Bayesian network to disambiguate between possible interpretations of the same speech utterance, given the context. The speech recognizer can return an N-best list of hypotheses, ranked by the acoustic likelihood. Our model provides a natural way of revising such ranking by incorporating information of the situation the robot is currently facing.

Similarly to Table~\ref{tab:useexamples}, Table~\ref{tab:useexamples2} shows a situation in which a number of objects are in the range of the robot's sensory inputs. As before, both results with the network trained on labeled speech data (a) and recognized speech data (b) are shown. The utterances corresponding to each column in the Table are, this time, the simulated output of a speech recognizer in the form of an N-best list with length three. The numbers below each hypothesis show the corresponding acoustic probability returned by the recognizer ($p(W_S^j)$ in Eq.~\ref{eq:combinerecog}). The other difference from Table~\ref{tab:useexamples} is that the probabilities in each entry are computed as in the bracketed expression in Eq.~\ref{eq:combinerecog}, i.e., by multiplying $p(w_i|X)$ for each word $w_i$ in the hypothesis. Finally, the final scores correspond to the full right term in Eq.~\ref{eq:combinerecog} summed over all available objects.

The probabilities in Table~\ref{tab:useexamples2}(a)~and~(b) are slightly different, but the result is the same: In both cases, the hypotheses of the recognizer are re-scored and the second hypothesis is selected when the posterior probability over all possible actions and objects is computed. Although this is just an illustrative example, it does suggest that, in spite of the less clean structure learned in the noisy conditions, the Bayesian network is still able to perform meaningful and useful inference.

\begin{figure*}
\includegraphics[width=\textwidth]{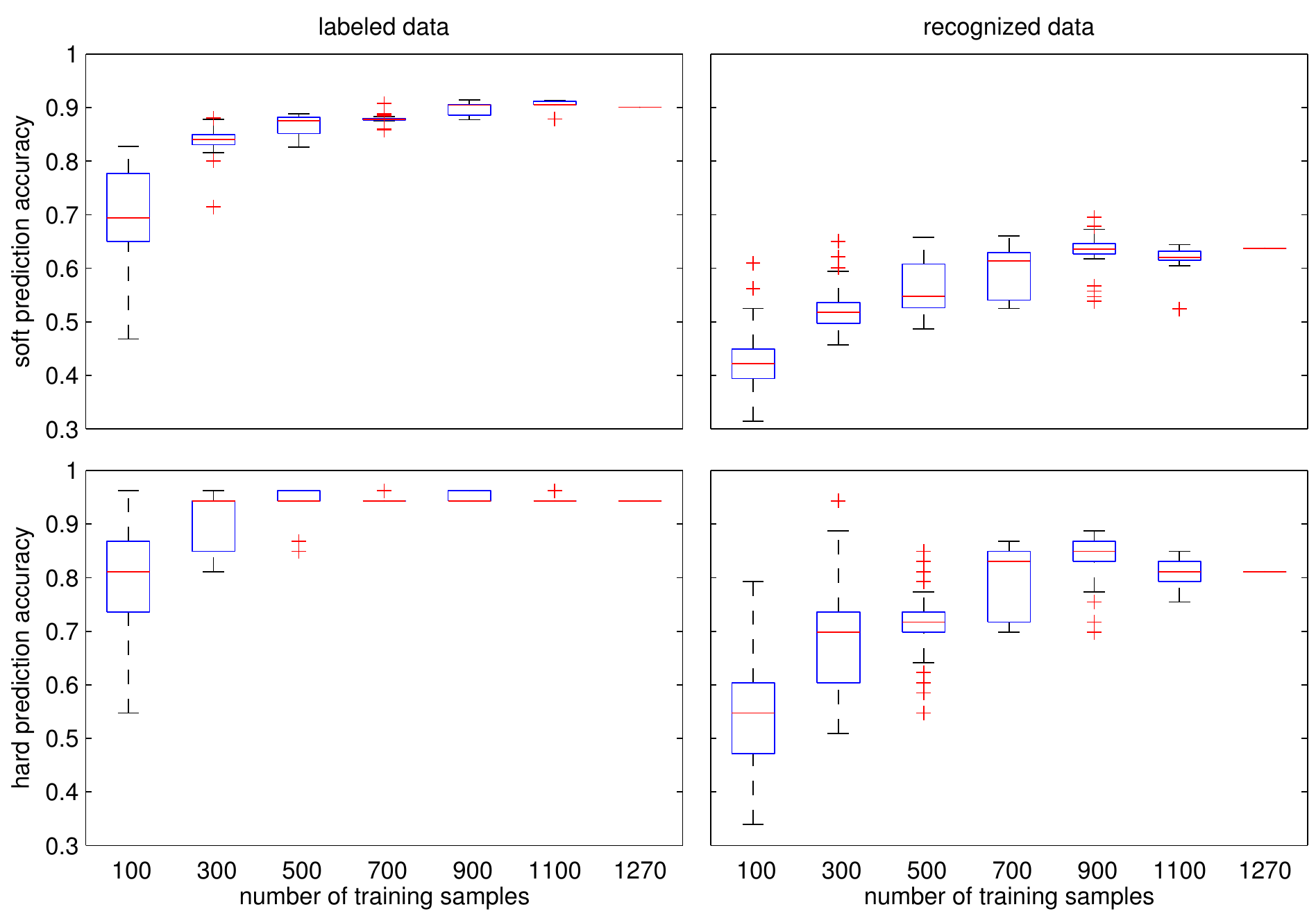}
%\vspace*{-5mm}
\caption{Staged learning with labeled (left) and recognized speech input (right). Top: soft prediction accuracy. Bottom: hard prediction accuracy}
\label{fig:stagedlearning}
\end{figure*}

\subsection{Quantitative Evaluation}
In order to evaluate the model in a quantitative way, the set of instructions described in Section~\ref{sec:data} was used. The task was to predict the object properties and the actions that are compatible with each, possibly incomplete, instruction.

Because the verbal instructions are often ambiguous, the right response is not unique. In order to score the model, we considered two scoring criteria. The first includes the response of the model for all the possibilities that were judged compatible with the verbal instruction by the human judges. We call this \emph{soft prediction accuracy}. In the second case, called \emph{hard prediction accuracy}, we consider only the best answer given by the model.

The soft prediction accuracy is computed in the following way: We calculate the marginal distribution of object properties and actions, given the verbal instruction. Then we sum the probabilities only over the object properties and actions that were considered correct by the human judges. The sum constitutes our measure of prediction accuracy for that particular example. If the model assigns nonzero probabilities to all and only the cases that are compatible with the verbal description, the sum is equal to 1. If the model gives nonzero probabilities to cases that are incompatible with the verbal instruction, the sum will be less that 1, and it will be closer to zero the more incompatible cases are favored by the model. These values are then averaged over the set of instructions in the test set.

For example, with the instruction ``move the small blue ball'', all object properties are unambiguous (color=blue, size=small, shape=sphere), but the action could both be grasp or tap. We, therefore, sum the probabilities we obtain from the network for both actions. If the only nonzero probabilities given by the model are for (color=blue, size=small, shape=sphere, action=grasp) and (color=blue, size=small, shape=sphere, action=tap), then the accuracy is 1, because the marginal distribution must sum to one. In any other case the accuracy will be less than 1.

 The hard prediction accuracy, more simply, counts the proportion of times the best prediction of the model is among the right cases specified by the human judges. This measure is more informative for the practical use of the model to control the robot's actions, but ignores the ability of the model to predict all the alternative correct answers.

In order to measure the effect of the affordance network on the results, we additionally trained a Bayesian network with no dependencies between the affordance nodes and where there is only a one-to-one relation between each word node and an affordance node.

The prediction results are summarized in Table~\ref{tab:summary}. A two way analysis of variance was also run separately for the soft and hard scores using recognition method (lab,asr) and network topology (with or without affordances) as independent variables. For the soft score we used a linear model and factorial analysis, whereas for the hard score a logistic model and $\chi^2$ test were used.

Firstly, we observe a degradation caused by recognition errors in all cases, effect that is significant with ($p=5.361\times 10^{-9} < 0.001$) for the soft prediction accuracy but not significant for the hard prediction accuracy, suggesting that the degradation might not be relevant in practical situations. Secondly, we can see that modeling affordances introduces a consistent and significant improvement (soft: $p=1.637\times 10^{-6}<0.001$, hard: $p=0.002<0.01$), compared to modeling only dependencies between words and either an action, an object visual property, or an effect. This can be explained by considering that the latter model is limited when it comes to disambiguating an incomplete instruction or detecting impossible requests. Also, as Figures~\ref{fig:verticalmovement}~and~\ref{fig:horizontalmovement} illustrate, effects are difficult to describe with single variables.

\begin{table}
\centering
\caption{Summary of prediction results}\label{tab:summary}
\begin{tabular}{l|cccc}
\hline\hline
training data       & \multicolumn{2}{c}{labeled} & \multicolumn{2}{c}{recognized} \\
prediction accuracy & soft       & hard        & soft        & hard \\ \hline
without affordances & 0.68       & 0.71        & 0.45        & 0.70 \\
with affordances    & 0.90       & 0.94        & 0.64        & 0.81 \\
\hline\hline
\end{tabular}
\end{table}

Another aspect that is interesting to measure is the dependency of our results with the amount of training data. In order to test this we trained the network with a varying number of training examples from 100 to 1270 with steps of 200. For each case the training was repeated 50 times (with randomly selected examples out of the 1270 total) and the corresponding network was tested on the instruction data set. The results are shown in Figure~\ref{fig:stagedlearning} both for labeled speech data and for recognized speech data with box-plots. The plots show the medians, quartiles and outliers of the soft and hard prediction accuracies for the 50 repetitions and for each stage. In the case of 1270 training examples, no variation can be seen, because there is only one way of selecting 1270 examples out of 1270. In most cases it can be seen that above 300 training examples, the prediction accuracy is relatively flat.

% do not remove (emacs configuration)
% Local variables:
% enable-local-variables: t
% ispell-local-dictionary: "british"
% mode: latex
% eval: (flyspell-mode)
% eval: (flyspell-buffer)
% End:

%\input{conclusions}
\section{Conclusions and Future Work}
\label{sec:conclusions}

This paper proposes a common framework to model affordances and to associate words to their meaning in a robotic manipulation task.  The model exploits co-occurrence between its own actions and a description provided by a human to infer the correct associations between words and actions, object properties and action's outcomes. Experimental results show that the robot is able to learn clear word-to-meaning association graphs from a set of 49 words and a dozen of concepts with just a few hundred human-robot-world interaction experiences. The learned associations were then used to instruct the robot and to include context information in the speech recognizer.

Although the structure learned in the noisy conditions given by the speech recognizer were somewhat less clean and interpretable, the model learned was still able to produce reasonable inference. A visible limit of the noisy model was the reduced ability to detect verbal inputs that were incompatible with the given situation, or even intrinsically. This is due to the fact that the model has learned out of sometimes inconsistent inputs caused by recognition errors. We believe that more training examples and an iterative learning, where the context learned so far is used to improve speech recognition, may solve this problem.

Based on these results, there are many extensions for our language acquisition model. On one hand, ongoing work on learning affordances will provide more complex models of the interaction of the robot with the environment \cite{Kjellstrom11,song2011}. This will open the door to learn larger sets of meanings in more complex and detailed situations.  
 On the other hand, we are currently investigating how to to relax some of our assumptions. In particular, we plan to include more complex robot-human interaction and social cues to allow a less rigid language between the instructor and the robot. Furthermore, it would be desirable to test if this model is able to predict some of the results that are observed with situational learning in early language acquisition experiments with human infants.

We believe that the encouraging results with our approach may afford robots with a capacity to acquire language descriptors in their operation's environment as well as to shed some light as to how this challenging process develops with human infants.

\bibliographystyle{IEEEtran}
\bibliography{references}

\begin{IEEEbiography}[{\includegraphics[width=1in,height=1.25in,clip,keepaspectratio]{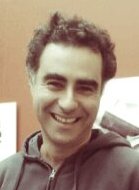}}]{Giampiero Salvi}
received the MSc degree in Electrical Engineering in 1998 from Università la Sapienza (Rome, Italy) and the PhD degree in Computer Science in 2006 from Kungliga Tekniska Högskolan (KTH, Stockholm, Sweden). From 2007 to 2009, he was a post-doctoral fellow at the Institute of Systems and Robotics (ISR), Lisbon. He is currently assistant professor and researcher at the department for Speech, Music and Hearing, School of Computer Science and Communication (KTH). He participates in several national and international research projects in the areas of speech technology, cognitive systems and robotics. He published several articles in international journals and conferences and his research interests include machine learning and speech technology.
\end{IEEEbiography}

\begin{IEEEbiography}[{\includegraphics[width=1in,height=1.25in,clip,keepaspectratio]{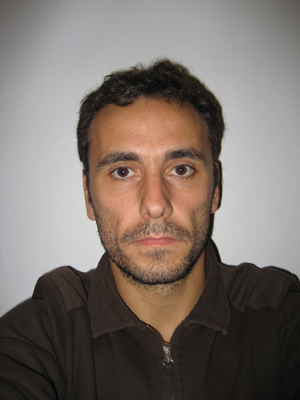}}]{Luis Montesano}
received his Ph.D. degree in computer science in 2006 from the University of Zaragoza, Spain. From 2006 to 2009, he was a Researcher at the Institute of Systems and Robotics (ISR), Lisbon. He is currently an assistant professor at the Computer Science Department of the University of Zaragoza, Spain. He has participated in various international research projects in the areas of mobile robotics and cognitive systems. His research interests include robotics, and machine learning.
\end{IEEEbiography}

\begin{IEEEbiography}[{\includegraphics[width=1in,height=1.25in,clip,keepaspectratio]{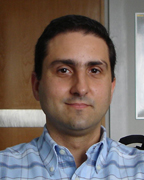}}]{Alexandre Bernardino}
received the PhD degree in Electrical and Computer Engineering in 2004 from Instituto Superior Técnico (IST). He is an Assistant Professor at IST and Researcher at the Institute for Systems and Robotics (ISR-Lisboa) in the Computer Vision Laboratory (VisLab). He participates in several national and international research projects in the fields of robotics, cognitive systems, computer vision and surveillance. He published several articles in international journals and conferences, and his main research interests focus on the application of computer vision, cognitive science and control theory to advanced robotics and automation systems.

Prof. Alexandre Bernardino is an IEEE member since 2006.
\end{IEEEbiography}

\begin{IEEEbiography}[{\includegraphics[width=1in,height=1.25in,clip,keepaspectratio]{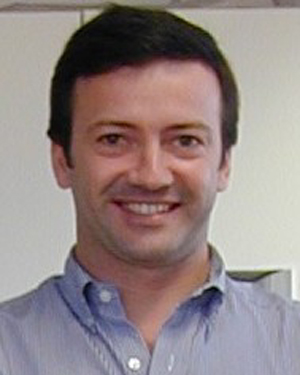}}]{José Santos-Victor}
received the PhD degree in Electrical and Computer Engineering in 1995 from Instituto Superior Técnico (IST - Lisbon, Portugal), in the area of Computer Vision and Robotics. He is an Associate Professor at the Department of Electrical and Computer Engineering of IST and a researcher of the Institute of Systems and Robotics (ISR), at the Computer and Robot Vision Lab - VisLab.  (http://vislab.isr.ist.utl.pt)

He is the scientific responsible for the participation of IST in various European and National research projects in the areas of Computer Vision and Robotics. His research interests are in the areas of Computer and Robot Vision, particularly in the relationship between visual perception and the control of action, biologically inspired vision and robotics, cognitive vision and visual controlled (land, air and underwater) mobile robots. 

Prof. Santos-Victor is an IEEE member since 1985 and an Associated Editor of the IEEE Transactions on Robotics.
\end{IEEEbiography}

% do not remove (emacs configuration)
% Local variables:
% enable-local-variables: t
% ispell-local-dictionary: "british"
% mode: latex
% eval: (flyspell-mode)
% eval: (flyspell-buffer)
% End:

\end{document}